\documentclass{article}
\usepackage[top=1in, bottom=1in, left=1in, right=1in]{geometry} 
\usepackage{graphicx} 
\usepackage{hyperref} 
\usepackage{subcaption} 
\usepackage{caption} 
\usepackage[style=apa, backend=biber]{biblatex}
\DeclareLanguageMapping{english}{english-apa}
\usepackage{tcolorbox} 
\usepackage{amsmath} 

\title{Vector Ontologies as an LLM world view extraction method}
\author{Kaspar Rothenfusser, Bekk Blando}
\date{February 2025}

\begin{document}

\maketitle

\section{Abstract}

Large Language Models (LLMs) possess intricate internal representations of the world, yet these latent structures are notoriously difficult to interpret or repurpose beyond the original prediction task. Building on our earlier work \parencite{Paper0}, which introduced the concept of vector ontologies as a framework for translating high-dimensional neural representations into interpretable geometric structures, this paper provides the first empirical validation of that approach. A vector ontology defines a domain-specific vector space spanned by ontologically meaningful dimensions, allowing geometric analysis of concepts and relationships within a domain. We construct an 8-dimensional vector ontology of musical genres based on Spotify audio features and test whether an LLM’s internal world model of music can be consistently and accurately projected into this space. Using GPT-4o-mini, we extract genre representations through multiple natural language prompts and analyze the consistency of these projections across linguistic variations and their alignment with ground-truth data. Our results show (1) high spatial consistency of genre projections across 47 query formulations, (2) strong alignment between LLM-inferred genre locations and real-world audio feature distributions, and (3) evidence of a direct relationship between prompt phrasing and spatial shifts in the LLM’s inferred vector ontology. These findings demonstrate that LLMs internalize structured, repurposable knowledge and that vector ontologies offer a promising method for extracting and analyzing this knowledge in a transparent and verifiable way.
\\
The code for this paper can be found at \textit{https://github.com/Thoughtful-Oasis/LLM-Worldview-Extraction-using-vector-ontologies}

\section{Introduction}

Vector representations and vector spaces are deeply connected to artificial intelligence and deep learning. Artificial neural networks (ANNs) rely almost entirely on vector manipulations to perform tasks. These vectors are very high-dimensional representations of the data, often understood to span feature spaces where each dimension represents a specific feature of the data (e.g., a patient's blood pressure, height, and weight).  However, in reality, these feature spaces are not as interpretable as this explanation suggests but rather span an abstract high-dimensional feature embedding without a clear mapping to real-world meaning. 

Some work has been done on this interpretability problem, most prominently by Anthropic \parencite{lindsey2025biology}. Anthropic's technique uses a replacement model trained to create a mapping from the LLM's hidden state onto an interpretable sparse model. This results in a kind of logic tree, or circuit, as they call it, allowing for tracing multi-step reasoning. Anthropic has also published work on neuron superposition within LLMs, which proposes that neurons can represent or relate to multiple features \parencite{elhage2022superposition}. They handle this in the replacement model approach by manually grouping neurons or by splitting features. 

Our approach differs in several ways. Anthropic's approach of using sparse models to interpret LLMs assumes that a neuron represents a singular feature/concept. Accordingly, superposition is proposed by Anthropic as a method of the model to compress many features into a limited feature space.

In contrast, our methodology assumes that neuron activations translate within or transform the feature space. Meaning that a neuron necessarily relates to numerous concepts of the LLM's worldview. In addition, we do not modify or require modifying the LLM, and we have less focus on creating a circuit structure and more focus on creating a geometric mapping between the latent space and the interpretable low-dimensional space we give the LLM.

Deep learning has shown great success in many tasks and acts as a general framework to learn world models of many kinds to analyze image, text, and numeric data distributions. Despite, the research above, the world models generated in deep learning lack effective interpretability. It is for that reason these world models can not be used for tasks other than the type of prediction task that generated them. This means that even though a model learns complex truths about its training data, these truths cannot be extracted and repurposed with current methods. 

Ontologies, on the other hand, are a world model type widely used in software and computer science. Ontological systems such as knowledge graphs apply a rigid top-down structure to the data \parencite{cimiano2016knowledge}. While this is interpretable, it has clear limitations since it necessarily fails to handle the various nuances that could be inductively learned from the data and hence is not necessarily reflective of the empirical truth \parencite{cimiano2016knowledge}. However, a NN approach has the inverse problems, mainly that it is not considered interpretable, but it is able to handle detailed nuances and is an inherently inductive approach to creating a statistical world model of the data.

In this paper, we test several hypotheses following from \cite{Paper0} by successfully utilizing vector ontologies to extract the learned statistical world model of LLMs into an ontological framework that remains interpretable and native to both neural networks and humans.

\section{Vector Ontologies}

A vector ontology of a given domain D, as proposed in \cite{Paper0}, is defined as a Vector space spanned by a finite set of $n$ ontological dimensions or basis vectors. These dimensions represent properties that are interpretable and ontologically meaningful to the domain.  For example, in a domain of apples, ontological dimensions might include sweetness, juiciness,  and crispness. Adopting the notation used in \cite{Paper0}, we can formally write the definition of the vector ontology of Apples as follows:\\
\[
V_{\text{ont}_{\text{apples}}} = \text{span} \{ x_1, x_2, x_3 \}, \quad x_i \in B_{\text{apples}}, 
\]
\[
\text{where} \quad B_{\text{apples}} = \{ \text{sweetness}, \text{juiciness}, \text{crispness} \}.
\]
These domain dimensions (basis vectors) represent scales that a domain entity displays to a larger or smaller degree, representing (mostly) continuous measures. Accordingly, an entity is defined by its position in the vector space spanned by those dimensions.\\
\[
\text {Granny Smith}=[0.2, 0.5, 0.8] \in V_{ont_{apples}}
\]

\[
\text {Fuji}=[0.8, 0.7, 0.7] \in V_{ont_{apples}}
\]

By constraining the ontology to this geometric form, ontological concepts are remapped onto geometric ones. Relationships become direction vectors, groups, and types and categories become subspaces and high-dimensional regions, and properties become positions in the space \parencite{Paper0}. 

This vector space represents something similar to a formal ontology, meaning that it constrains possible realities to all points within the space. However, in reality, of course, not all points in the space are equally likely to appear, due to correlations between these domain dimensions (e.g., weight and size of an apple strongly correlate).
\\

In \parencite{Paper0} we suggest that Artificial intelligence systems use vector ontologies to perform tasks. Analyzing this hypothesis, we conducted a series of experiments to extract the world model of a pre-trained large language model into a vector ontology and compare it to a ground truth ontology of identical structure.

\section{World Model Extraction and Verification}

Given the theoretical and empirical evidence that LLMs contain a high-resolution internal world model of the domain they were trained on, we hypothesize that this model can be extracted from the model's latent space and projected onto a predefined Vector ontology. We further hypothesize that this projected world model effectively maps onto the true (ground truth) data distribution in the same vector ontology, formalizing the following two hypotheses.
\\
\\
\textbf{Hypothesis 1:} A part of the learned internal world model of Large Language Models can be extracted from its latent space by mapping it onto a predefined Vector ontology (through token prediction). This mapping is close to deterministic (i.e., reasonably stable despite token prediction limitations).
\\
\\
\textbf{Hypothesis 2:} This mapped world model is accurate, meaning that it effectively maps onto the true data distribution in an identical vector ontology populated with ground truth data.
\\
\\
This means that, for example, in a domain of apples, we expect to extract the model's internal idea of which apple lies in which point of a 2d vector ontology:
\[
V_{apples} = \{x_1: \text{sweetness},  
 x_2:\text{juiciness}\}
\]

We expect this mapping to be highly stable, meaning that the LLM places Granny Smith consistently at $[0.2, 0.5]$ and not suddenly at $[0.9, 0.8]$ when we use different words to ask.  This stability indicates that we truly perform something like a projection or extraction from latent space and not just random text generation. Secondly, we expect the extracted position to correlate with the true placement generated with ground truth methods (e.g., human categorization). If this is the case, we argue that we have successfully extracted the model's internal worldview and shown that it is correct and hence potentially meaningfully re-purposable for other tasks.

\section{Methodology}
To test the above hypothesis, we utilize a Dataset of 15 million pre-labeled songs and test genre placement inside an 8-dimensional vector ontology.

\subsection{Dataset}
We extracted metadata for approximately 15 million songs using the Spotify web API. The extracted audio data includes values of each song's \textit{danceability, energy, speechiness, acousticness, instrumentalness, liveness, valence,} and \textit{tempo}, normalized to a range between 0 and 1 (tempo in BPM). The exact origin of these values is not publicly known, but according to the publicly available information, these ratings are generated using a mix of signal processing and a machine-learning model trained on human-labeled data. We hence argue they represent a sort of ground truth rating, as it approximates human perception indirectly and contains much domain knowledge from Spotify and originally Echonest.

This Data was used to create a vector ontology of Music with the above-mentioned 8 dimensions represented in formal notation as:

\[
V_{\text{ont}_{\text{songs}}} = \text{span} \{ x_1, x_2, x_3, ...,x_8 \}, \quad x_i \in B_{\text{songs}}, 
\]
\[
\text{where} \quad B_{\text{songs}} = 
\left\{ 
\begin{array}{ll}
    x_1 &= \text{danceability}, \quad x_2 = \text{energy}, \quad x_3 = \text{speechiness}, \quad x_4 = \text{acousticness}, \\
    x_5 &= \text{instrumentalness}, \quad x_6 = \text{liveness}, \quad x_7 = \text{valence}, \quad x_8 = \text{tempo}
\end{array}
\right\}.
\]

This eight-dimensional feature space represents a vector ontology of a Song Domain specifically targeted to ontologically classify a song regarding its "vibe".  Since we are interested in the distribution of points across the space, we discretized them using a high-dimensional binning strategy. This strategy divides each dimension into n ranges ${r_0,...,r_{n-1}}$ determined such that each range contains an equal number of data points. To force density, we iteratively increased n to maximize resolution while ensuring at least 50\% density, meaning that for at least half of the hypercubes formed by slicing the space at the range boundaries of each dimension, at least one datapoint (song) falls inside that hypercube. This strategy resulted in a final transformation of the continuous space into a discrete one found at $n=6$. This resulted in a total of $6^8 = 1 679 616 \approx 1.7  Million$ hypercubes representing bins (possible positions), of which 52.4\% were populated. Each point was placed in the appropriate bin and hence assigned a new position represented by the center point of the bin's hypercube.\\
This means that each song was assigned a discrete position in the space as an 8 dimensional vector containing integers between 0 and 4, indexing the bins following the schema shown in \ref{tab:binEdges} For example Beethovens 5th symphony which in the original continuous space is represented as $[0.3, 0.2, 0.0, 0.9, 0.9,0.1,0.2,95]$ was converted to $[0,0,0,4,5,1,1,1]$.

\begin{table}
    \centering
    \begin{tabular}{c|c|c|c|c|c|c}
         Dimension & $r_0$ & $r_1$ & $r_2$ & $r_3$ & $r_4$ & $r_5$ \\ \hline
         Danceability & {\scriptsize 0.00-0.35} & {\scriptsize 0.35-0.48} & {\scriptsize 0.48-0.58} & {\scriptsize 0.58-0.67} & {\scriptsize 0.67-0.76} & {\scriptsize 0.76-1.00} \\ \hline
         Energy & {\scriptsize 0.00-0.20} & {\scriptsize 0.20-0.42} & {\scriptsize 0.42-0.59} & {\scriptsize 0.59-0.73} & {\scriptsize 0.73-0.86} & {\scriptsize 0.86-1.00} \\ \hline
         Speechiness & {\scriptsize 0.00-0.03} & {\scriptsize 0.03-0.04} & {\scriptsize 0.04-0.05} & {\scriptsize 0.05-0.07} & {\scriptsize 0.07-0.13} & {\scriptsize 0.13-0.97} \\ \hline
         Acousticness & {\scriptsize 0.00-0.01} & {\scriptsize 0.01-0.06} & {\scriptsize 0.06-0.27} & {\scriptsize 0.27-0.66} & {\scriptsize 0.66-0.92} & {\scriptsize 0.92-1.00} \\ \hline
         Instrumentalness & {\scriptsize 0.00-0.00} & {\scriptsize 0.00-0.00} & {\scriptsize 0.00-0.03} & {\scriptsize 0.03-0.61} & {\scriptsize 0.61-0.87} & {\scriptsize 0.87-1.00} \\ \hline
         Liveness & {\scriptsize 0.00-0.08} & {\scriptsize 0.08-0.10} & {\scriptsize 0.10-0.12} & {\scriptsize 0.12-0.18} & {\scriptsize 0.18-0.33} & {\scriptsize 0.33-1.00} \\ \hline
         Valence & {\scriptsize 0.00-0.14} & {\scriptsize 0.14-0.29} & {\scriptsize 0.29-0.43} & {\scriptsize 0.43-0.59} & {\scriptsize 0.59-0.77} & {\scriptsize 0.77-1.00} \\ \hline
         Tempo & {\scriptsize 0.00-88}& {\scriptsize 88-105}& {\scriptsize 105-120}& {\scriptsize 120-129}& {\scriptsize 129-145}& {\scriptsize 145-250}\\ \hline
    \end{tabular}
    \caption{Vector space discretization ranges}
    \label{tab:binEdges}
\end{table}

\subsection{LLM world model extraction}

The Large Language Model used in this experiment was the GPT-4o-mini model provided through the openai api client. This model was chosen due to its good performance at a low cost. No fine-tuning or additional instructions other than the ones stated below were used. 

To extract the internal world model of the LLM and specifically to project it onto the vector ontology explained above, the model was informed of the vector ontology's structure and tasked to identify positions in that space corresponding to user queries. The returned position was then used to retrieve songs falling in the high-dimensional bin at that position using our database.
\\
\\
\begin{tcolorbox}[colback=lightgray!5!lightgray, colframe=black!75!black, title=LLM prompting]
\textbf{System Prompt:}
\texttt{
Answer in a single JSON object with an entry called 'location', which is a list
of variables formatted as dimension names as keys and indices as values. 
These describe the location in the 8D vibe space, with the dimensions
[danceability', 'energy', 'speechiness', 'acousticness', 'instrumentalness', 'liveness', 'valence', and 'tempo'] taken by a vibe corresponding
to the music you would recommend based on the chat history. Return variables as indices corresponding to the bucket values in this pattern (zero-indexed) ensuring that the values are within the range of the bins (0-5): <FeatureRanges>}\footnote{represents a dictionary with the above table as a list of objects formatted as: \{'name': 'danceability', 'ranges': ['0.0-0.35', '0.35-0.48', '0.48-0.58', '0.58-0.67', '0.67-76', '0.7-1.0]\}}
\\
\\
\textbf{Example completion:}\\
\textit{User:} I have a headache\\
\textit{System:} \texttt{[0,0,0,3,4,1,0,0]} $\rightarrow$ sample song: (\href{https://open.spotify.com/track/6lFF792Pd4bDwF9kanGpFx?si=c59943a463c34d5f}{Great Blue by K. Okabayashi})\\
\end{tcolorbox}
\textbf{Genre Analysis}
\\
To test the extent to which the model has a consistent world model in this setup and to what degree that model is consistent with the true data distribution, it was tasked to identify the position of 50 different genres in the vector ontology. Since a key concern when utilizing LLMs is their probabilistic nature, we introduced a measure of reliability and model consistency by using 47 different query formulations for each genre (examples below). The goal of this was the creation of semantic variance in the internal vector representation of the prompt while keeping the task performed constant (i.e., locating the same genre). The returned positions were saved for further analysis, acting as the LLM world model projection into our predefined vector ontology.
\\

Query formulation examples:
\begin{itemize}
    \item Direct: "\texttt{<genre>}"
    \item Emotional state: "I'm feeling \texttt{<genre>}"
    \item Mood-based: "I'm in the mood for \texttt{<genre>} music"
    \item Action-oriented: 
    \begin{itemize}
        \item "Find me \texttt{<genre>} music"
        \item "Play me \texttt{<genre>} songs"Preference: "I want \texttt{<genre>} music"
    \end{itemize}
\end{itemize}

\subsection{Ground Truth World Model}
In addition to the world model extracted from the LLM, we create a reference world model acting as the ground truth. Similarly to the model extracted from the LLM, we focus on the distributions of genres across the space.

To compare a genre's location as determined by the LLM to its true position, we determine both the genre distribution found at the LLM-specified positions and the true location of genres (or their distribution) inside the vector ontology acting as the ground truth. We determined the genre of each song using the Spotify API. \\

Unfortunately, Spotify doesn't provide direct genre information on a song. Therefore, each song's artist was retrieved, and afterward, for each of those artists, a list of their most fitting genres was compiled. The set of all genres across the artists of a song was then added to a counter, keeping track of the total genre count in the specific bin (location). Although this doesn't perfectly indicate the genre of the location, as artist and song genres might not always be equivalent, we assume it to have a strong correlation, and across all the songs per location, it should still yield meaningful genre assignment. Genre was calculated this way for every song in the dataset.

This reprocessing created detailed data of each position's genre distribution as well as a genre's occurrence across all 1,679,616 locations.

\section{Results}

\subsection{World model visualisation}

\begin{figure}
    \centering
    \includegraphics[width=0.9\linewidth]{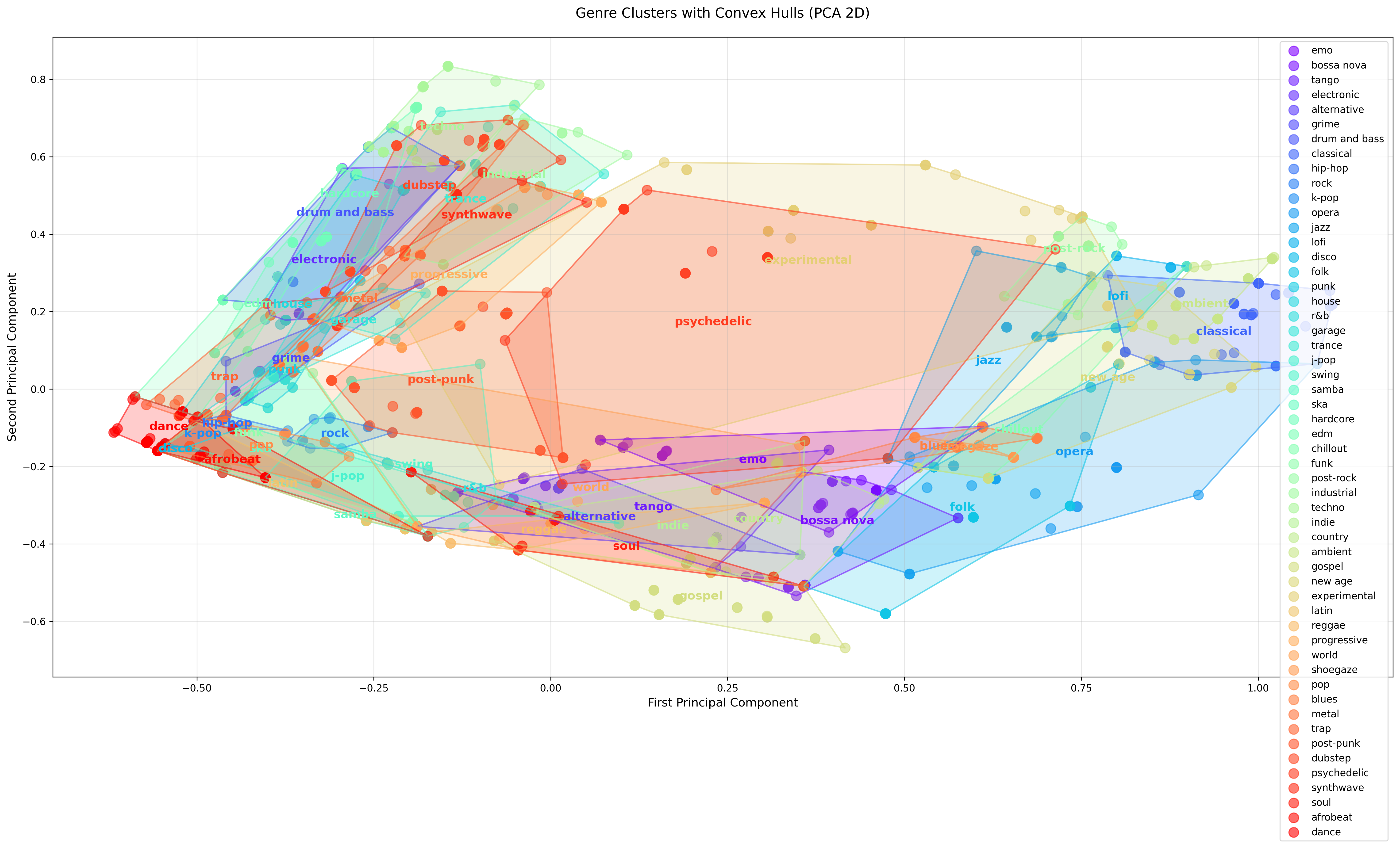}
    \caption{LLM extracted genre locations across query variations projected using PCA}
    \label{fig:genreClusters}
\end{figure}

To get an intuitive perspective of the world model extracted from the LLM, we performed a principal component analysis (PCA) to visually plot genre search locations in clusters. First, we performed PCA on all individual points created with the different query formulations across all genres. In this manner, we reduced the 8-dimensional audio feature space to 2D and plotted convex hulls for each genre to show their search boundaries (Figure \ref{fig:genreClusters}). This visualization reveals the full spread and potential overlap of genre searches. Second, we calculated the average position (centroid) of each genre's searches in the original 8D space and visualized these midpoints using the same PCA transformation to create a slightly cleaner visualization containing only average genre position without spread.

\begin{figure}
    \centering
    \includegraphics[width=0.9\linewidth]{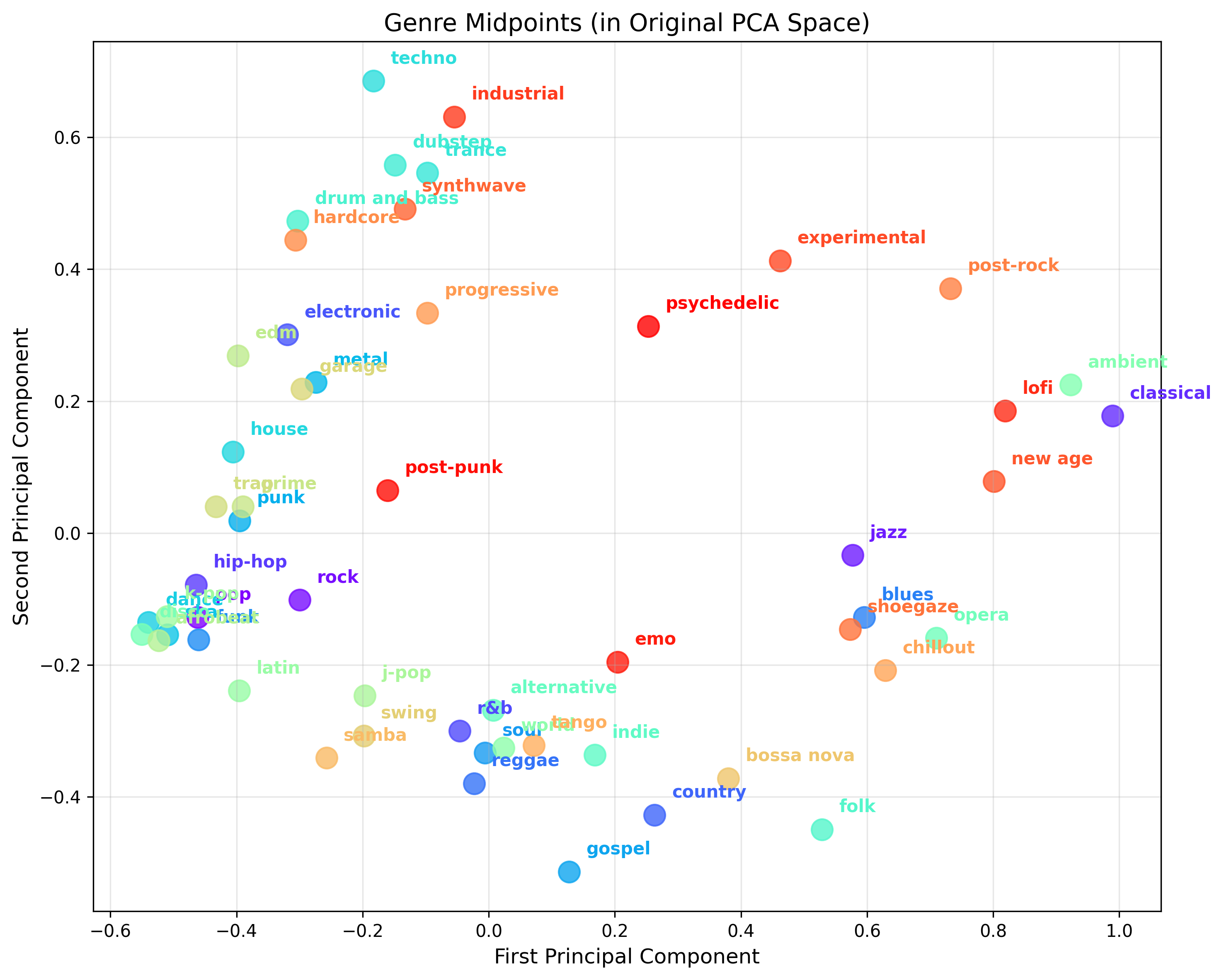}
    \caption{LLM extracted genre locations as the centroid of query variations projected using PCA}
    \label{fig:genreCentroids}
\end{figure}

Figures \ref{fig:genreClusters} and \ref{fig:genreCentroids} clearly show that the models spatially separate genres while overlapping between several of them. This is to be expected, given that the audio features alone do not fully separate genres (see next sections). We find, however, that the spatial distributions of genres, as well as the overlap between genre regions as perceived by the model, seem to be in line with human understanding of music. For example, there is an overlap between classical and opera, dubstep and techno, and jazz and blues, but no overlap between classical and metal or techno and jazz.

Overall, it seems the model extracted music ontology incorporates a clear and reasonable understanding of the music universe.

\subsection{Consistency}

To measure the consistency of the world model extracted from the LLM as described above, we analyze the spatial distribution of search points for each genre across the 47 query formulations.
The consistency is evaluated through multiple metrics: (1) Total unique search locations across query-variations, (2) average distances of query-variation points from their centroid, (3) pairwise distance between query-variations, (4) hull dimensionality and volume, (5) percentage of space volume covered by search points.

To contextualize these metrics, we compare them against a random baseline generated by uniformly sampling 47 points from the possible locations in the discrete space as described in Table \ref{tab:binEdges}.

\subsubsection{Unique search locations}

\begin{figure}
    \centering
        \includegraphics[width=0.9\linewidth]{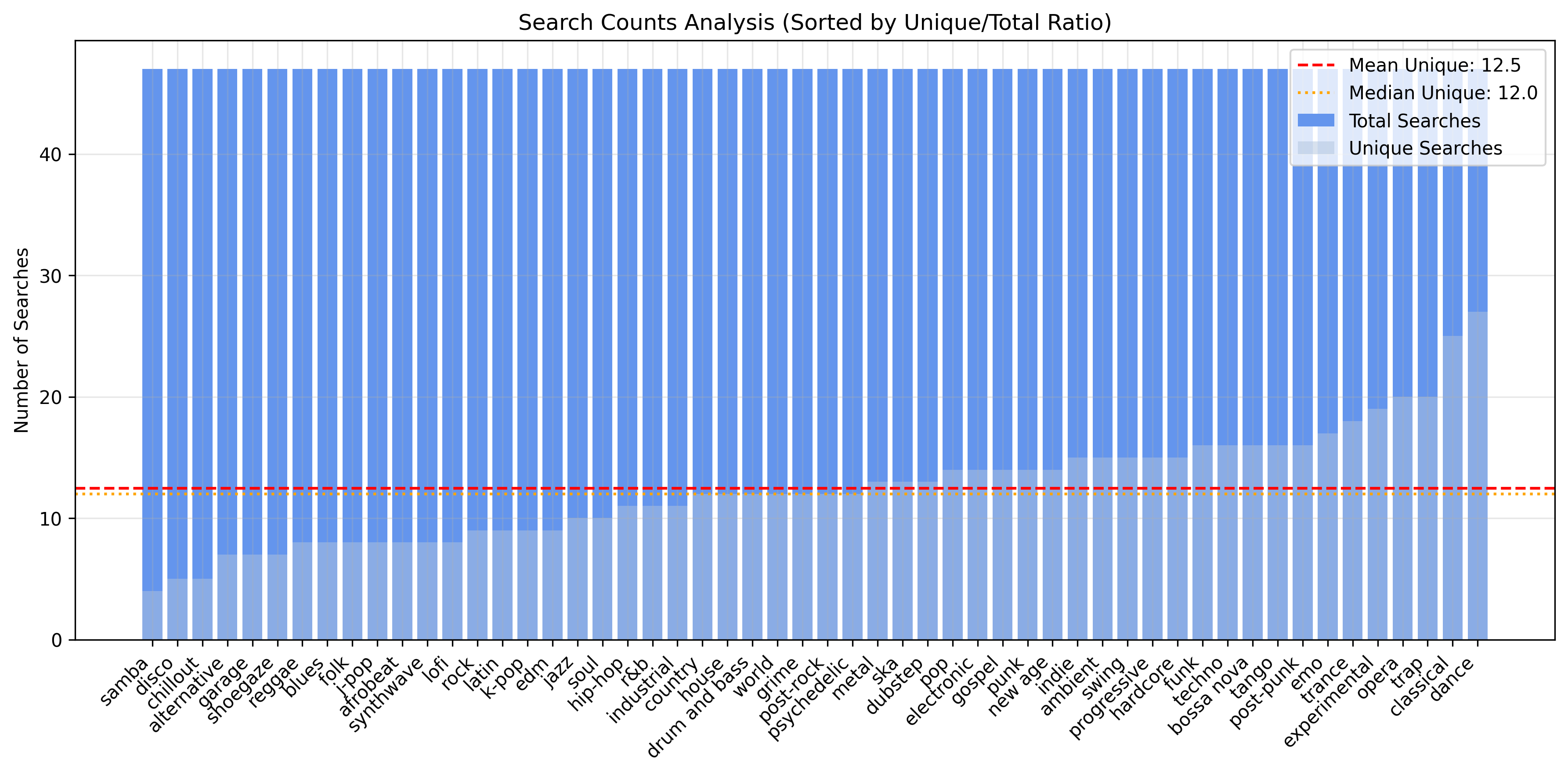}
        \caption{total and unique successful query location counts by genre}
        \label{fig:searchCount}
\end{figure}

The most direct and straightforward measure of consistency is the number of unique locations across query formulations. Figure \ref{fig:searchCount} shows that we retrieved 47 locations per genre (one per query formulation), which, on average, resulted in 12.5 unique positions. Some genres got as few as 4 unique locations across 47 generations, which is remarkably little given a total of 1.7 million possible locations.

\subsubsection{Centroid distance and pairwise Distance}

For a more nuanced analysis, we investigate two different Euclidean distance measures between the individual locations extracted from the LLM across query formulations for the same genre. Firstly, we calculate the centroid of all points for the same genre across query formulations. We then investigate the Euclidean distance between individual query formulations and their genre centroid, as well as the average pairwise distance between different query formulations for the same genre. \\
In an ideal, perfectly consistent worldview extraction, we expect the identical location to be returned by the LLM across all query formulations for the same genre (e.g., "Jazz", "Play some Jazz", "I want jazz", etc., all return the same coordinate). Accordingly, an average centroid distance of 0 and an average pairwise distance of 0 across all genres would indicate deterministic worldview extraction. The closer to fully inconsistent and hence unsuccessful the extraction is, however, the closer we expect the average centroid distance and pairwise distance to approach a random baseline, which samples groups of 47 random points in the space and calculates their average centroid and pairwise distance.
\begin{figure}
    \centering
        \includegraphics[width=0.9\linewidth]{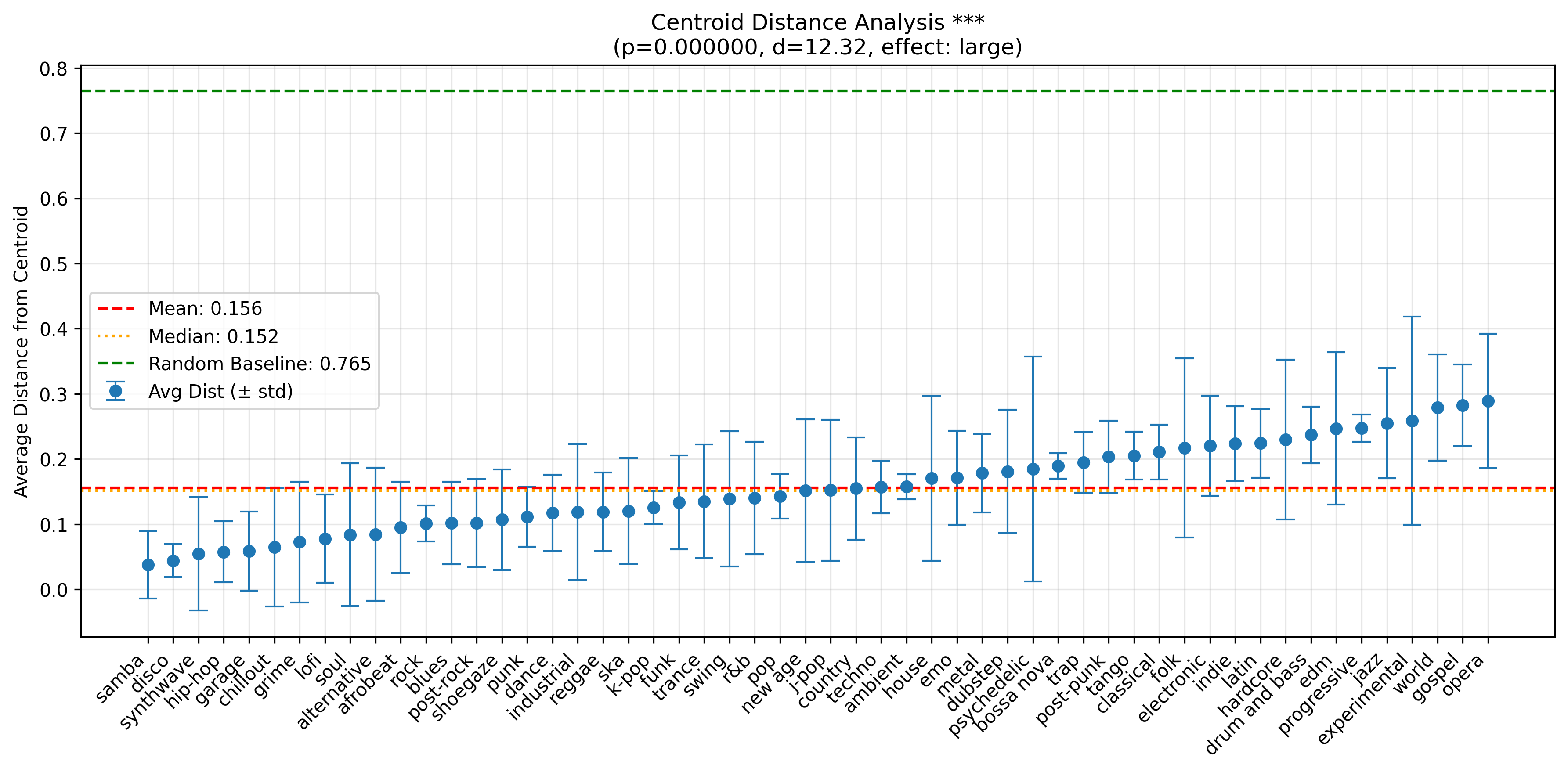}
        \caption{Euclidean Distance between individual query formulation for a genre and its centroid compared against a random baseline. Equal genre position across all query formulations would result in an average distance of 0, indicating perfect consistency. Higher distance hints towards less consistent genre positioning across query formulations.}
        \label{fig:centroidDistance}
\end{figure}

\begin{figure}
    \centering
    \includegraphics[width=0.9\linewidth]{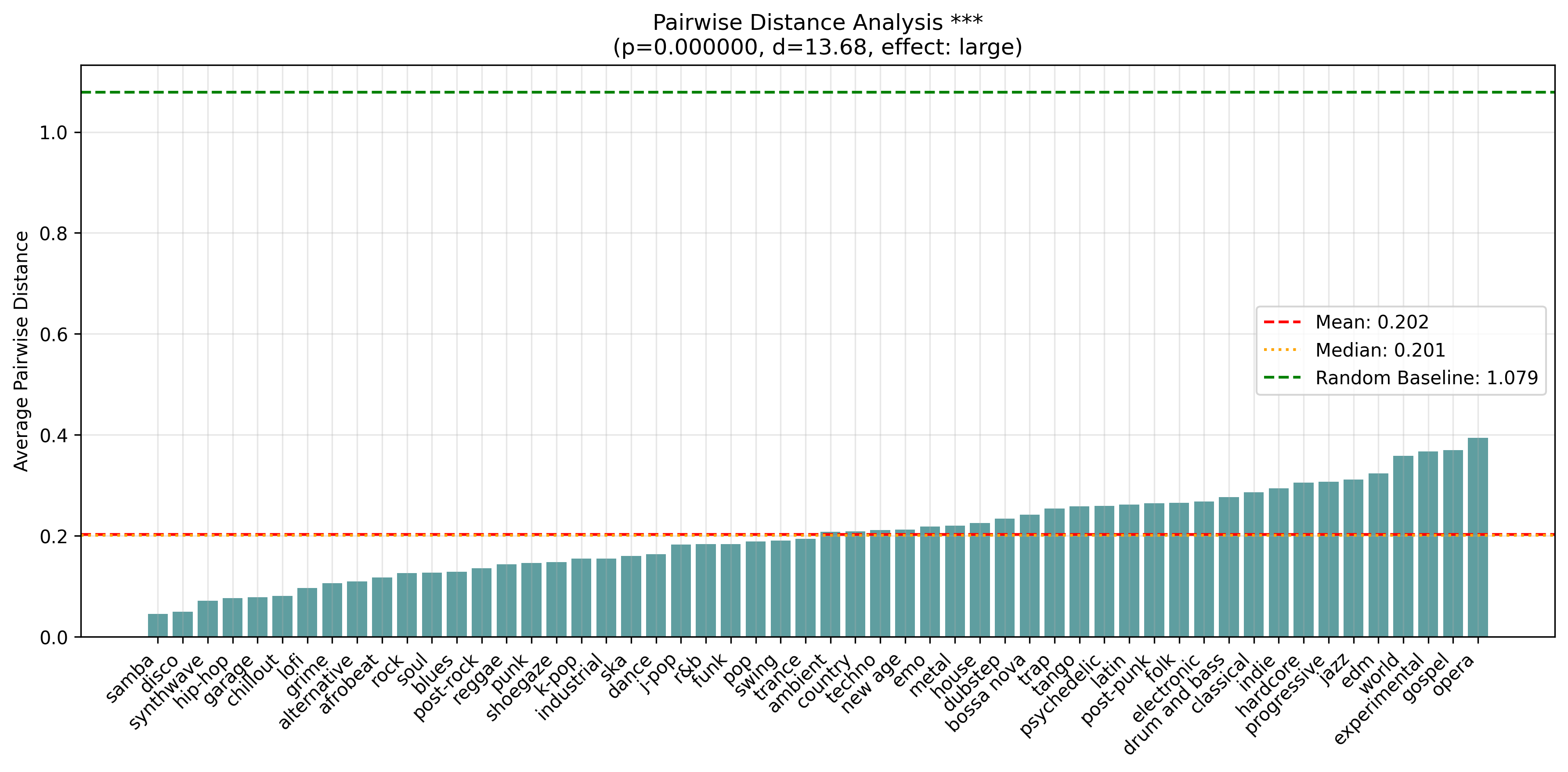}
        \caption{Average pairwise Euclidean distance between individual query formulations for the same genre compared against a random baseline}
        \label{fig:pairwiseDistance}
\end{figure}

Figure \ref{fig:centroidDistance} shows that the average centroid distance is very small compared to the random baseline. With a mean of 0.156 and a median of 0.152 across genres (vs. random: 0.765), these results show very high consistency with strong statistical significance (p \textless 0.000000, d=12.32). Figure \ref{fig:pairwiseDistance} shows even clearer results for the average pairwise distance, with a mean of 0.202 and median of 0.201 compared to a random baseline of 1.079 (p \textless 0.000000, d=13.68).\\
It is important to note that the consistency varies greatly across genres, with some genres being close to perfectly consistent \textit{(e.g., samba: centroid \textless 0.05, pairwise \textless  0.05)} while others being far less consistent, even though much better than random \textit{(e.g., opera: centroid \textless 0.3, pairwise \textless 0.4)}. \\

These results give further evidence for consistent worldview extraction. However, Euclidean distance doesn't capture the nuances of the geometry between the points, which is why the measures elaborated below give more concrete insights into the level of consistency.

\subsubsection{Subspace dimensionality}

\begin{figure}
    \centering
    \includegraphics[width=0.9\linewidth]{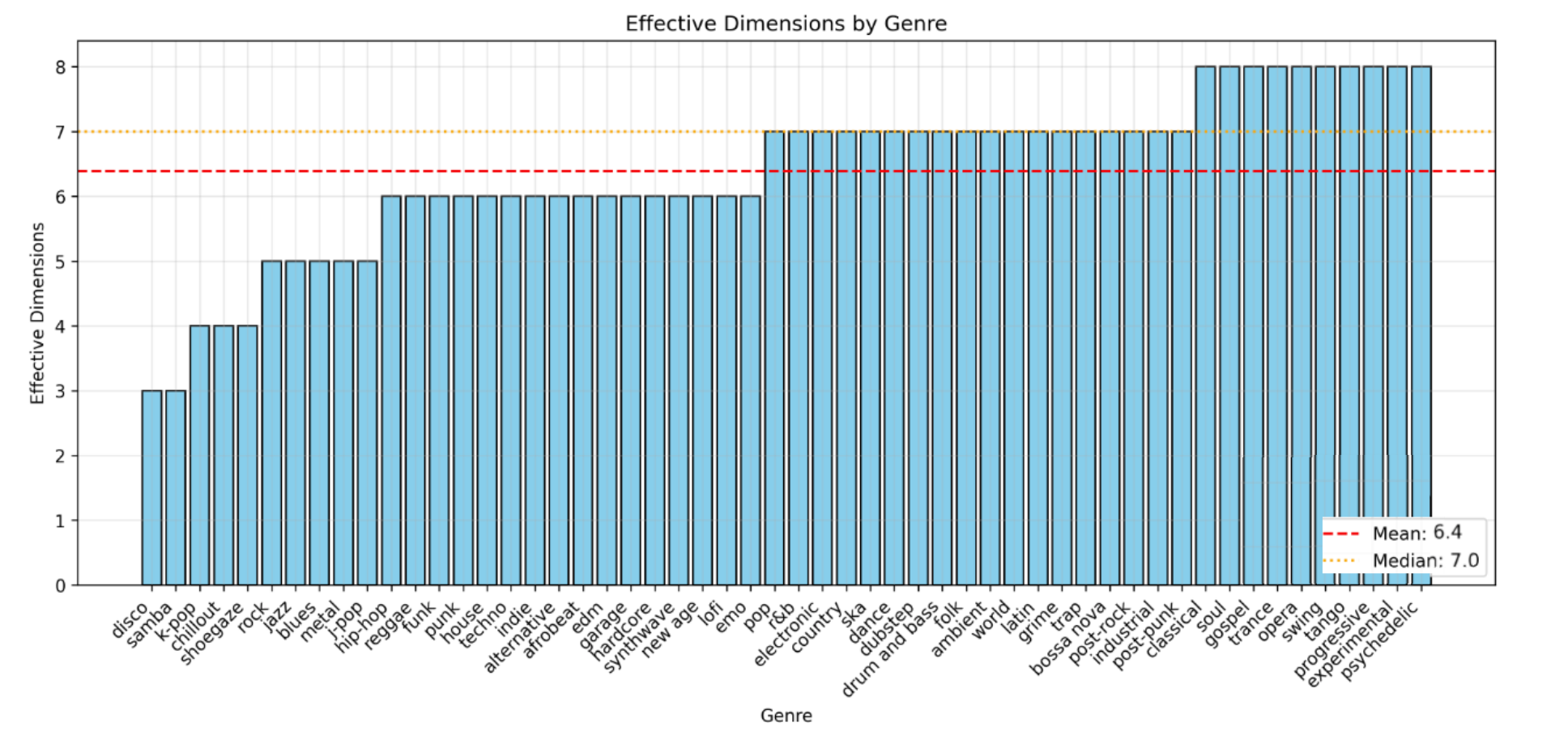}
        \caption{Dimensionality of hull connecting all individual query formulations for the same genre.}
        \label{fig:effectiveDimensions}
\end{figure}

To properly assess the consistency of the worldview extraction methodology performed, we need a more nuanced investigation than simple distances between points. This is because it seems reasonable to assume that a single genre might not occupy a single point in space but rather a geometric region. Accordingly, we might expect some variance in search location. However, we are interested in the structure of this variance. If, for example, the different query formulations create several locations, but all of these locations are situated in a 2d subspace (a plane), we might consider this to be more consistent and meaningful than points that are somewhat clustered but still occupy all 8 dimensions. We hence analyze the effective dimensionality of the variance between query formulations for the same genre, which is shown in figure \ref{fig:effectiveDimensions}.\\

We can see that most genres occupy a lower-dimensional subspace, with a mean of 6.4 dimensions. Some genres occupy as few as three dimensions. This indicates that the LLM's Worldview, when extracted in this manner, places genres in lower-dimensional manifolds (subspaces), which is in line with the manifold theory much discussed in AI research and further points towards stable worldview extraction.

\subsubsection{Percentage of space volume covered}

When dealing with high-dimensional spaces, distance measures become unintuitive and often lose their clear meaning. Accordingly, the distance measures investigated in Figure \ref{fig:centroidDistance} and \ref{fig:pairwiseDistance}, while meaningful, do not give us an intuitive perspective on consistency. For example, when looking at the centroid distance for the genre "opera", which performs worst of all genres at 0.3, we might compare it to the random baseline of 0.75 and assume a 60\% improvement. This, however, originates from the fact that distance is a one-dimensional measure. Our intuition tells us something like: "0.3 means that 30 percent of the points in the space are as close to the centroid as the search points, while for random, it's 75 percent". While this would be somewhat correct in one dimension, already in two dimensions, the equation becomes quadratic, and only $0.3^2 = 0.09 = 9\%$ of the points are as close or closer to the centroid than the search point. \\
Accordingly, for a more comprehensive understanding of search consistency, and one that comes closer to what we probably mean intuitively, we determine the percentage of space volume covered by the points. More concretely, we determine the volume of the hypersphere around the centroid with the radius of a) the mean and b) the maximum distance between individual query formulation search points and that centroid. We again compare this to a random baseline of 47 randomly sampled points. This gives us a volumetric percentage of the space covered by a genre's variance compared to the total space.\\

\begin{figure}
    \centering
    \includegraphics[width=0.85\linewidth]{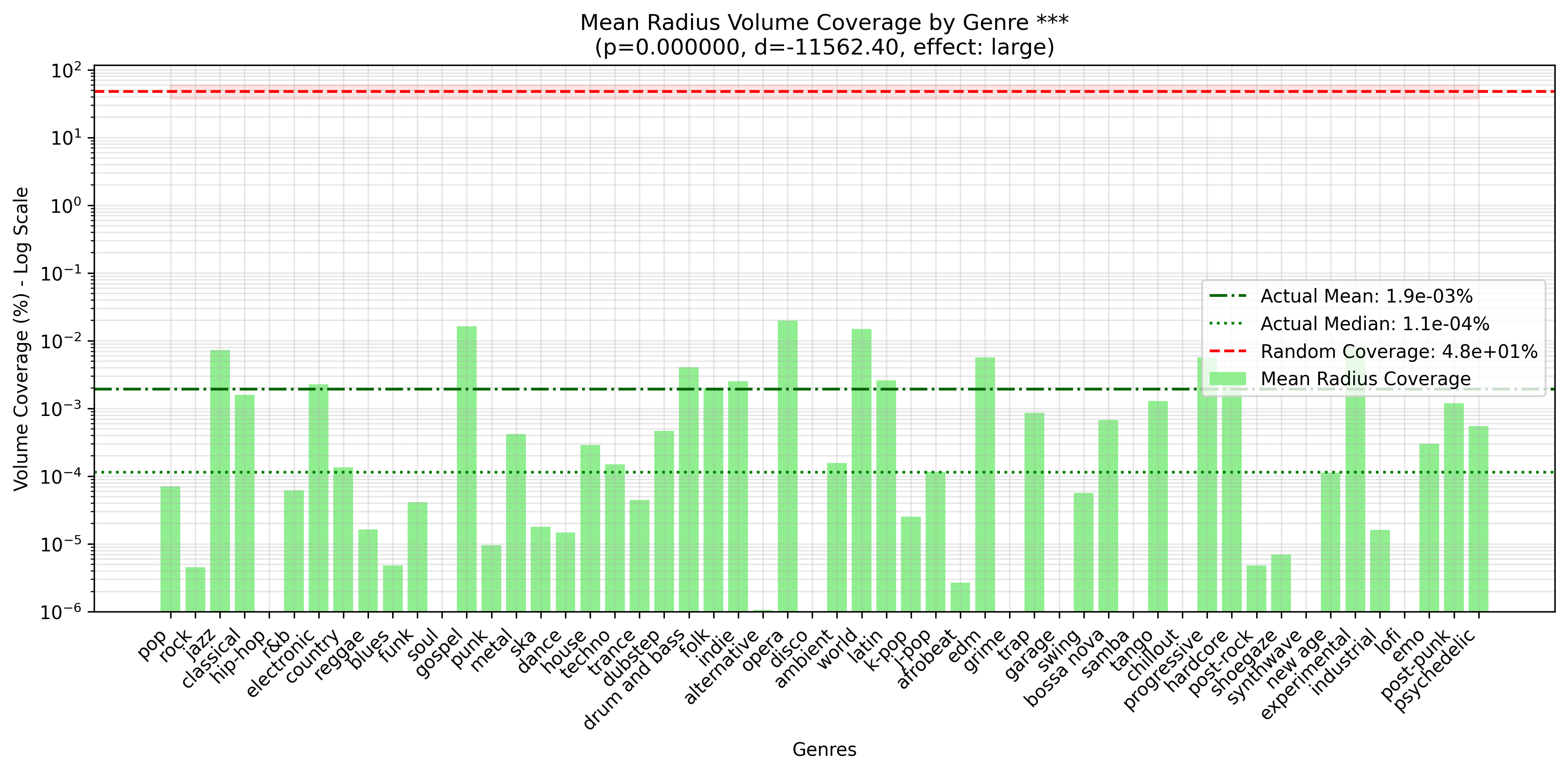}
        \caption{Volume of the hypersphere centered around the centroid of each genre with the radius of the genre's individual search point's mean centroid distance (as a percentage of the total space on a log scale)}
        \label{fig:volumeCoverageMean}
\end{figure}
\begin{figure}
    \centering
    \includegraphics[width=0.84\linewidth]{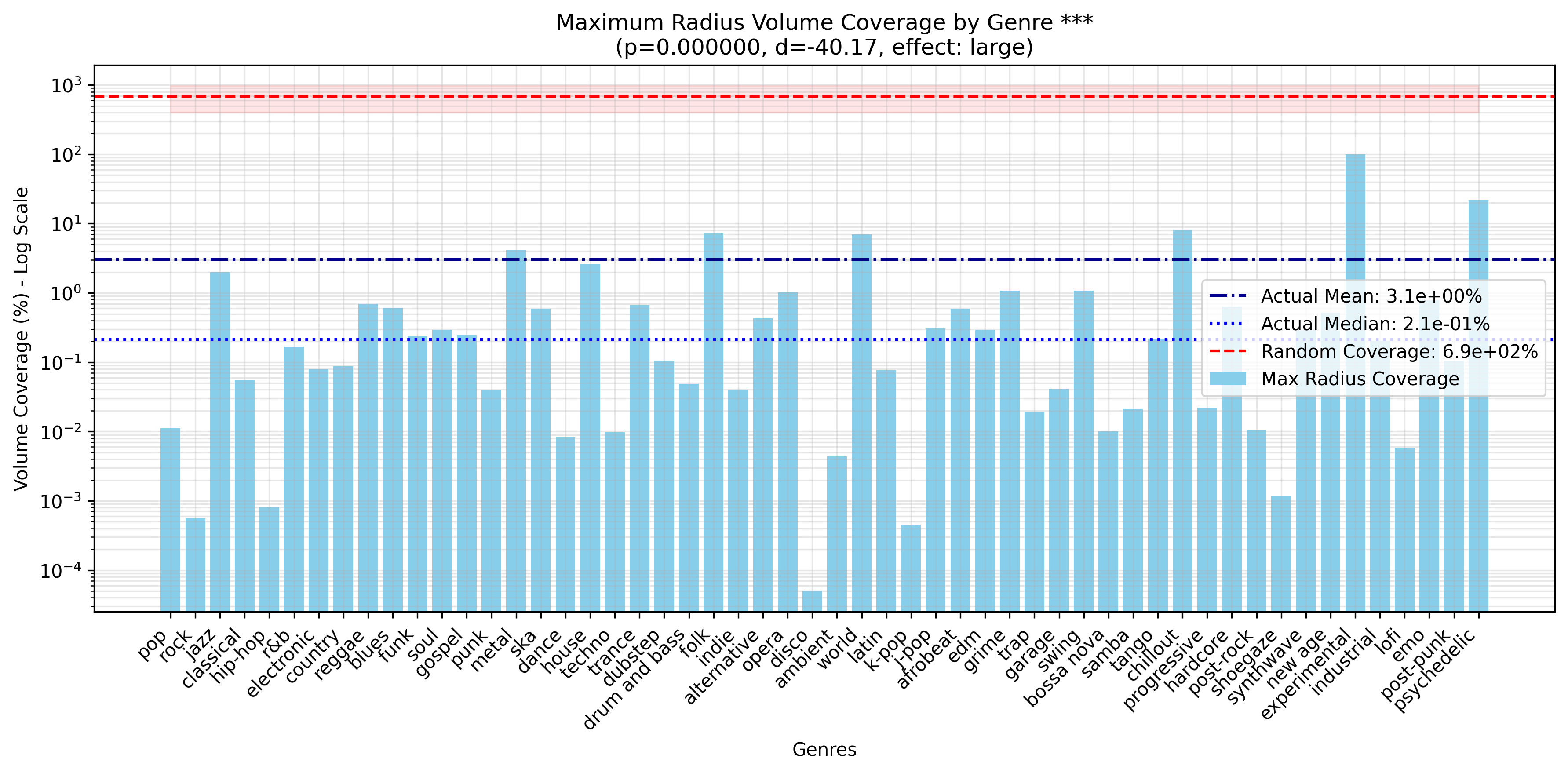}
        \caption{Volume of the hypersphere centered around the centroid of each genre with the radius of the genre's individual search point's max centroid distance (as a percentage of the total space on a log scale)}
        \label{fig:volumeCoverageMax}
\end{figure}

Figure \ref{fig:volumeCoverageMax} and Figure \ref{fig:volumeCoverageMean} visualize the extremely high consistency of the extracted worldview in terms of space volume coverage. When using the mean centroid distance, the mean percentage of space covered is 0.0019\% compared to 48\% in the random baseline ($>25.000$ times smaller), with the median significantly lower at 0.00011\% ($> 400.000$ times smaller than random). When using the more pessimistic max centroid distance, we still see an average space coverage of 3.1\%, with a median much lower at 0.21\%, which still is more than 200 and respectively 3.000 times smaller than the random baseline at 690\%\footnote{The volume larger than the total space for the random baseline originates in the fact that the radius sampled in this manner might be as large as 1, which creates a sphere that exceeds the space boundaries.}. Similar to the prior findings, we find the results to be highly significant and of large effect (p \textless0.000000, d = -11563 / -40).

This shows that across query formulations, the variance of extracted locations occupies a tiny fraction of the volumetric space. This supports Hypothesis 1 and demonstrates highly stable behavior of the extracted vector ontology.

\subsection{Accuracy}

While the above section clearly shows consistency in the LLMs world model when projected into our predefined vector ontology, the below section aims to analyze the accuracy of that world model when compared to our Ground Truth ontology.

To evaluate the accuracy of the LLM's understanding of musical genres, we employ three complementary analyses. First, we examine the distribution of genres found in the model's search locations for each genre, which shows both individual genre frequencies and broader genre group patterns, revealing how well the model's conception of a genre aligns with the actual genres found in the music database at the search location. Second, we visualize the PCA-reduced feature space through heatmaps of genre distributions overlaid with convex hulls of the model's search locations, providing a direct spatial comparison between where the model searches for a genre and where that genre actually appears in the musical feature space. Lastly, we quantitatively analyze the distance as well as the cosine similarity between the center points of genre searches and the distribution center points of the same genre in our ground truth ontology. The genre center points are calculated using the most significant genre hotspots and creating a weighted average of their locations based on genre presence.

\subsubsection{Distribution found}

\begin{figure*} 
    \centering

    \begin{subfigure}{0.24\textwidth} 
        \centering
        \includegraphics[width=\linewidth]{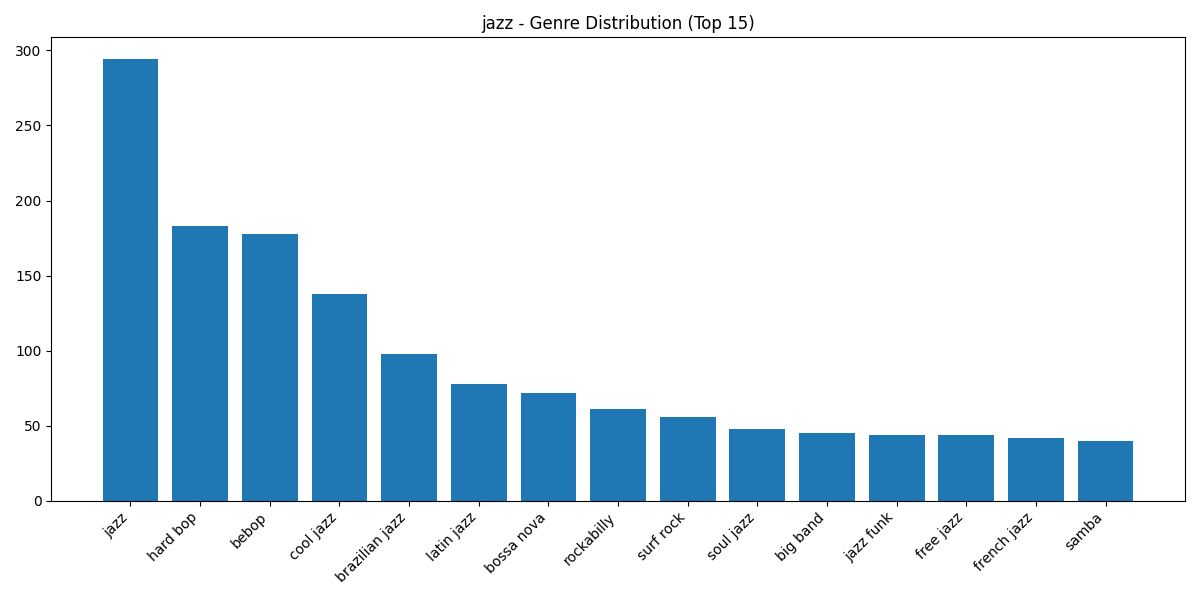}
        \caption{Jazz Distribution}
    \end{subfigure}
    \begin{subfigure}{0.24\textwidth}
        \centering
        \includegraphics[width=\linewidth]{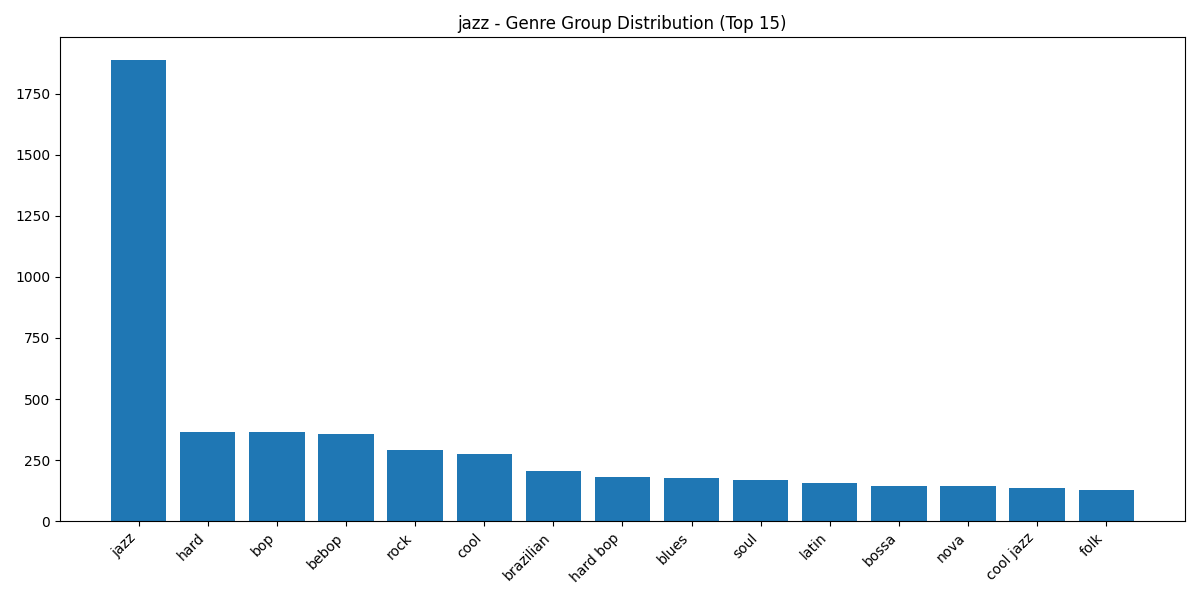}
        \caption{Jazz Group Distribution}
    \end{subfigure}
    \begin{subfigure}{0.24\textwidth}
        \centering
        \includegraphics[width=\linewidth]{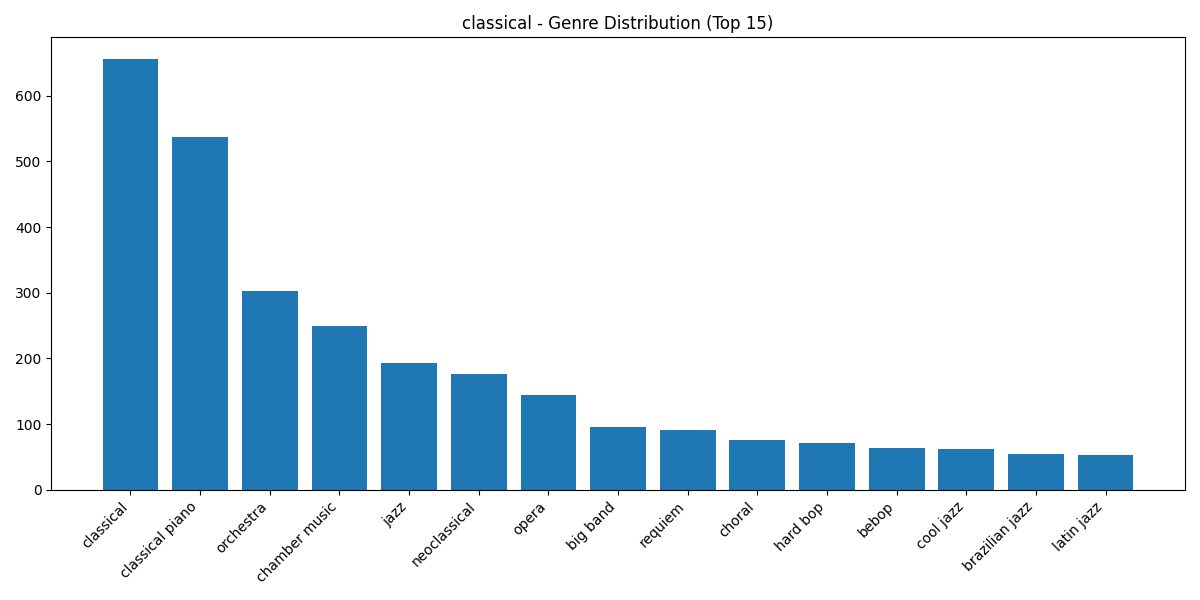}
        \caption{Classical Distribution}
    \end{subfigure}
    \begin{subfigure}{0.24\textwidth}
        \centering
        \includegraphics[width=\linewidth]{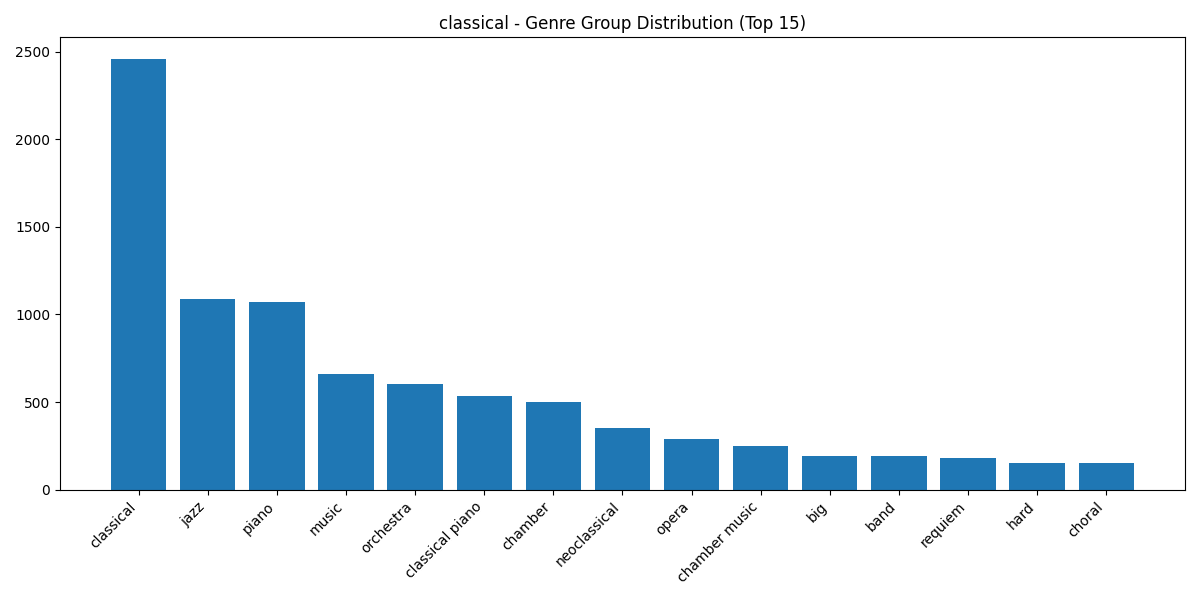}
        \caption{Classical Group Distribution}
    \end{subfigure}

    \begin{subfigure}{0.24\textwidth}
        \centering
        \includegraphics[width=\linewidth]{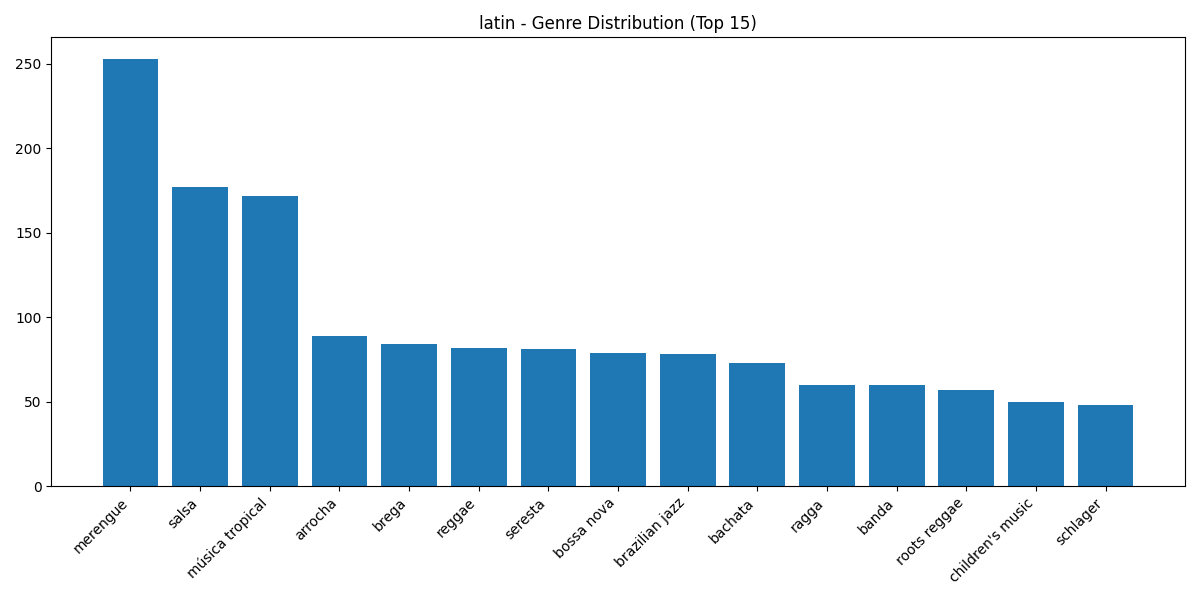}
        \caption{Latin Distribution}
    \end{subfigure}
    \begin{subfigure}{0.24\textwidth}
        \centering
        \includegraphics[width=\linewidth]{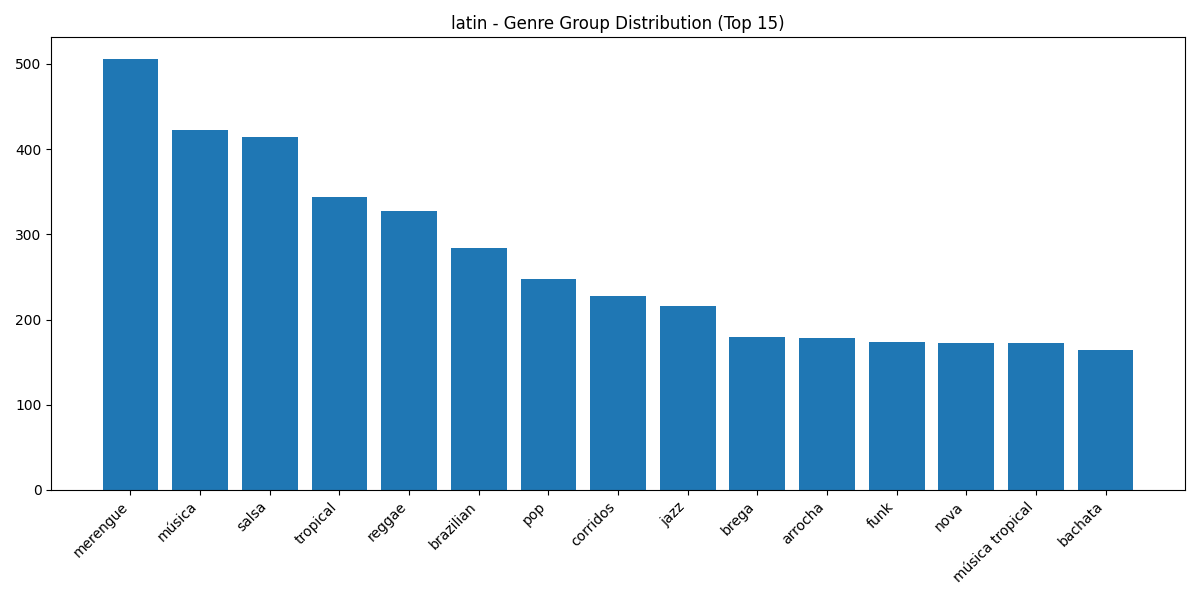}
        \caption{Latin Group Distribution}
    \end{subfigure}
    \begin{subfigure}{0.24\textwidth}
        \centering
        \includegraphics[width=\linewidth]{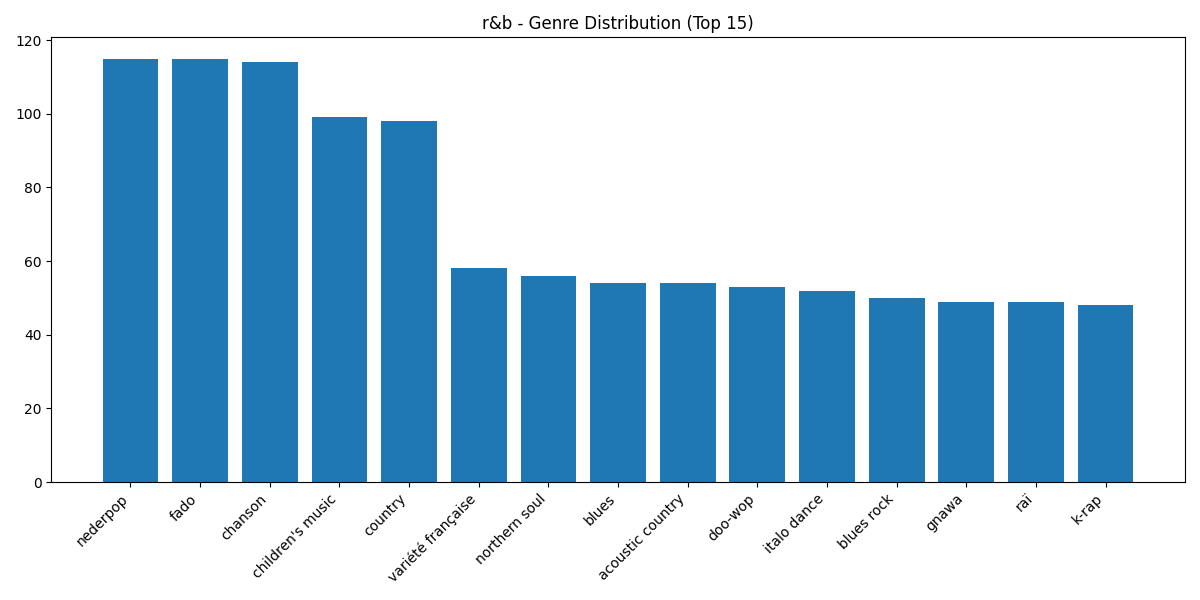}
        \caption{R\&B Distribution}
    \end{subfigure}
    \begin{subfigure}{0.24\textwidth}
        \centering
        \includegraphics[width=\linewidth]{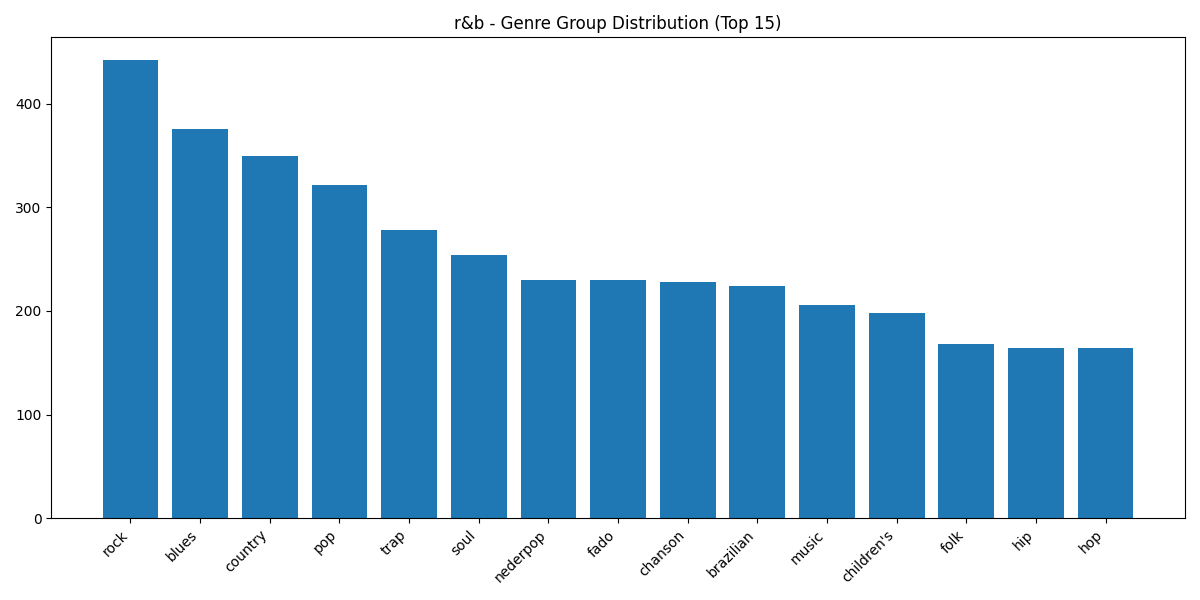}
        \caption{R\&B  Group Distribution}
    \end{subfigure}

    \begin{subfigure}{0.24\textwidth}
        \centering
        \includegraphics[width=\linewidth]{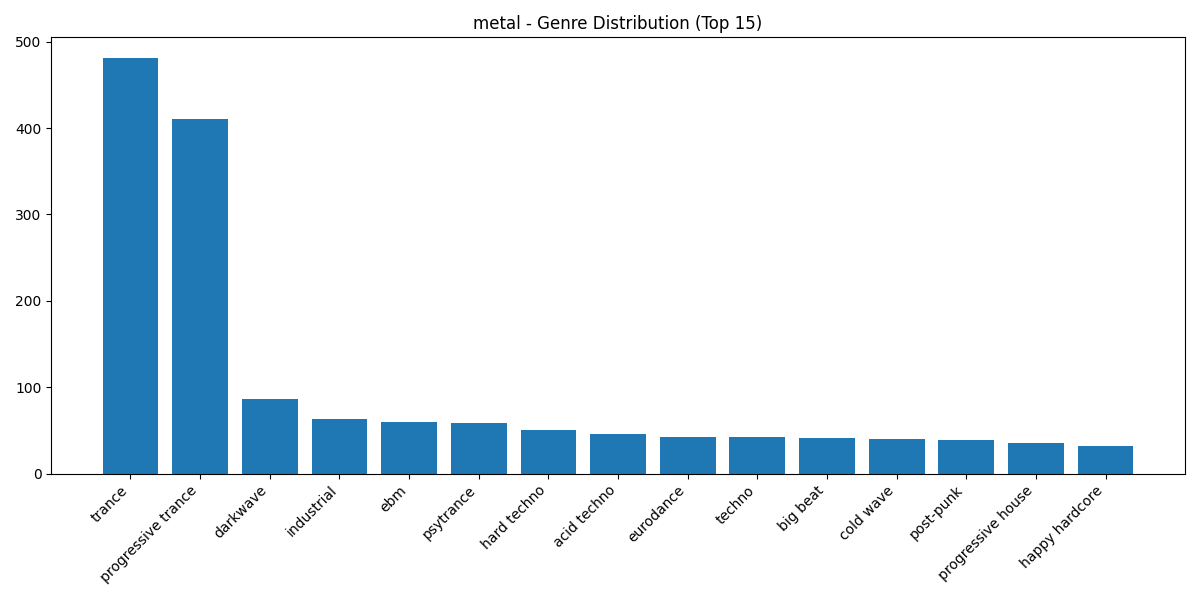}
        \caption{Metal Distribution}
    \end{subfigure}
    \begin{subfigure}{0.24\textwidth}
        \centering
        \includegraphics[width=\linewidth]{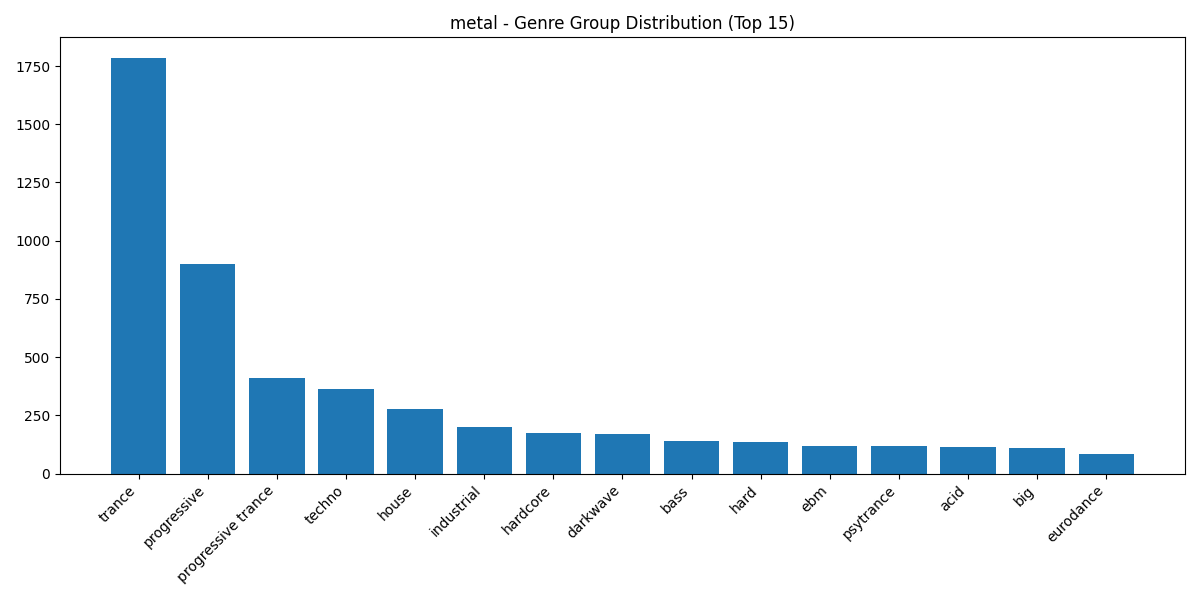}
        \caption{Metal Group Distribution}
    \end{subfigure}
    \begin{subfigure}{0.24\textwidth}
        \centering
        \includegraphics[width=\linewidth]{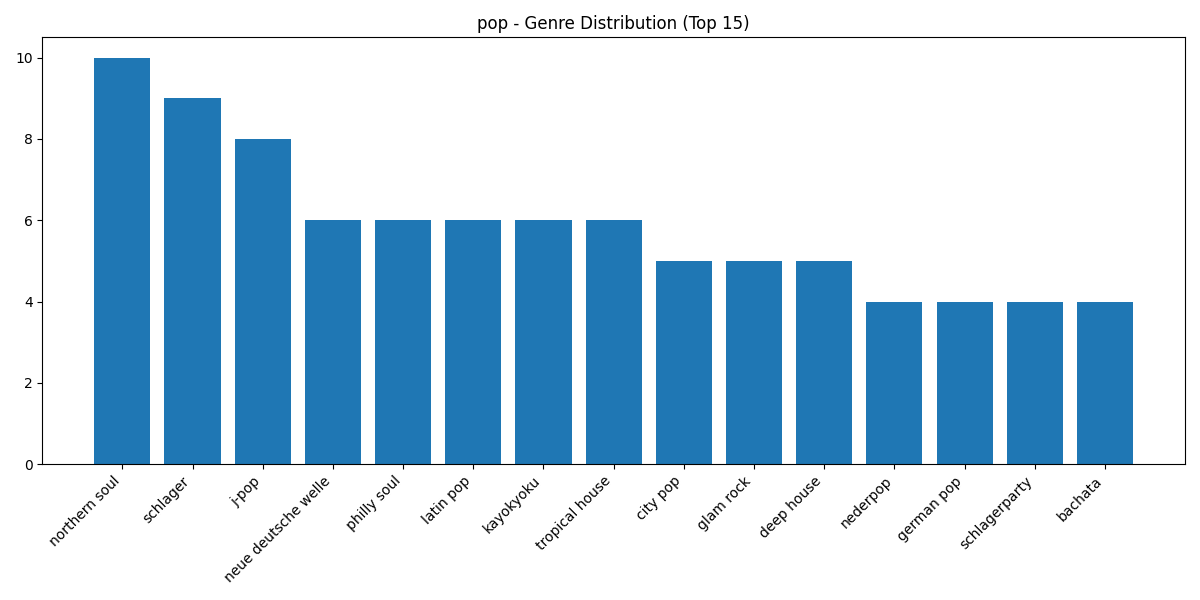}
        \caption{Pop Distribution}
    \end{subfigure}
    \begin{subfigure}{0.24\textwidth}
        \centering
        \includegraphics[width=\linewidth]{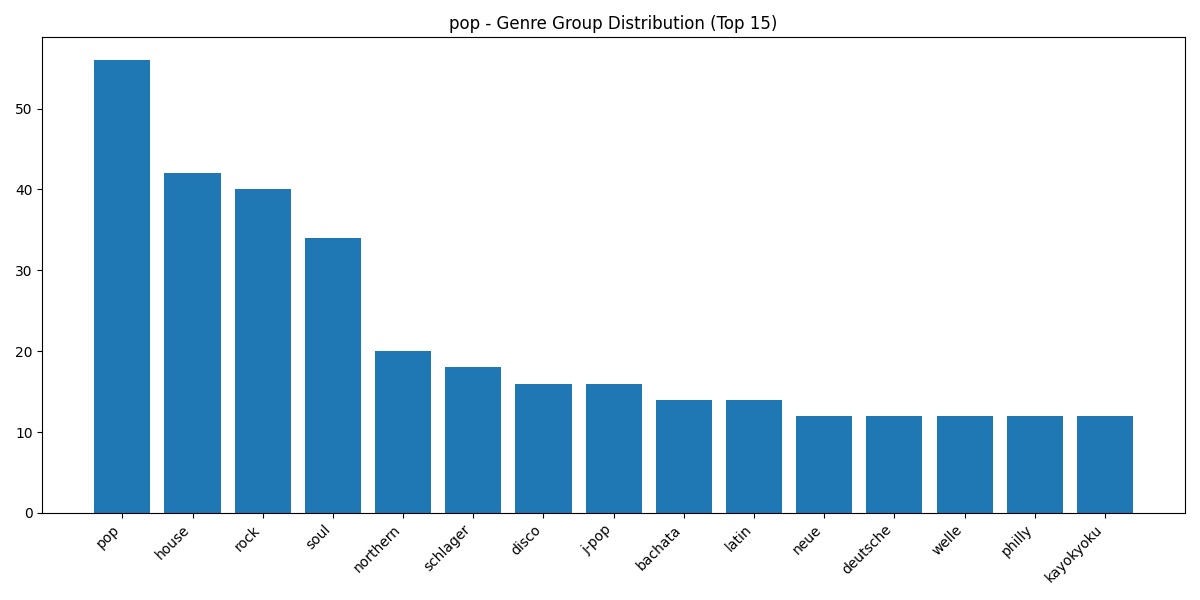}
        \caption{Pop Group Distribution}
    \end{subfigure}

    \begin{subfigure}{0.24\textwidth}
        \centering
        \includegraphics[width=\linewidth]{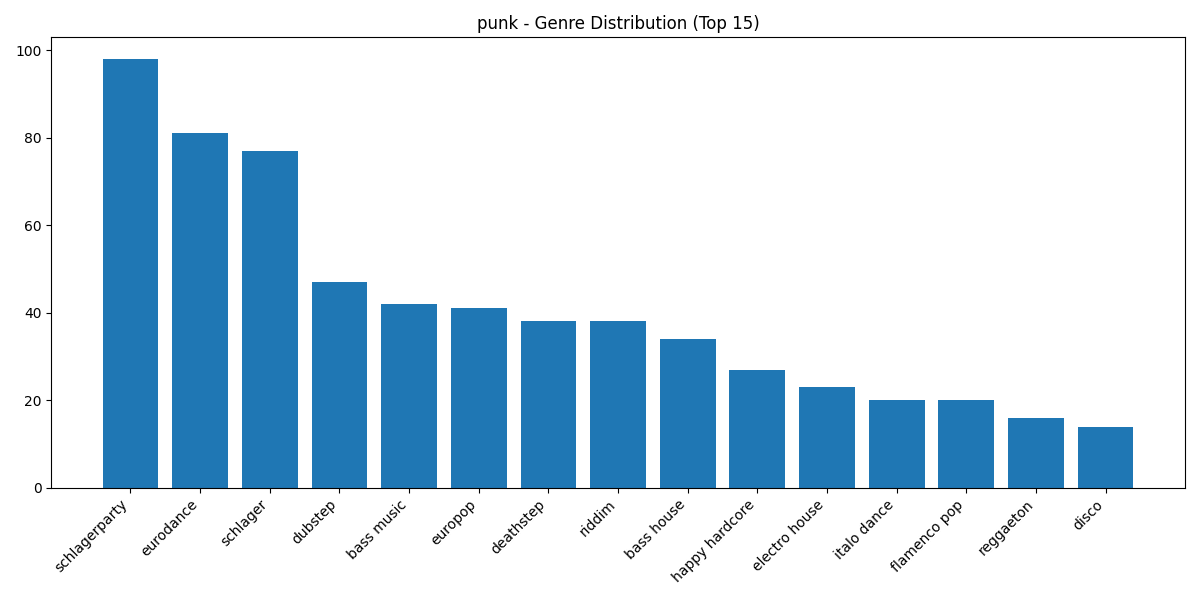}
        \caption{Punk Distribution}
    \end{subfigure}
    \begin{subfigure}{0.24\textwidth}
        \centering
        \includegraphics[width=\linewidth]{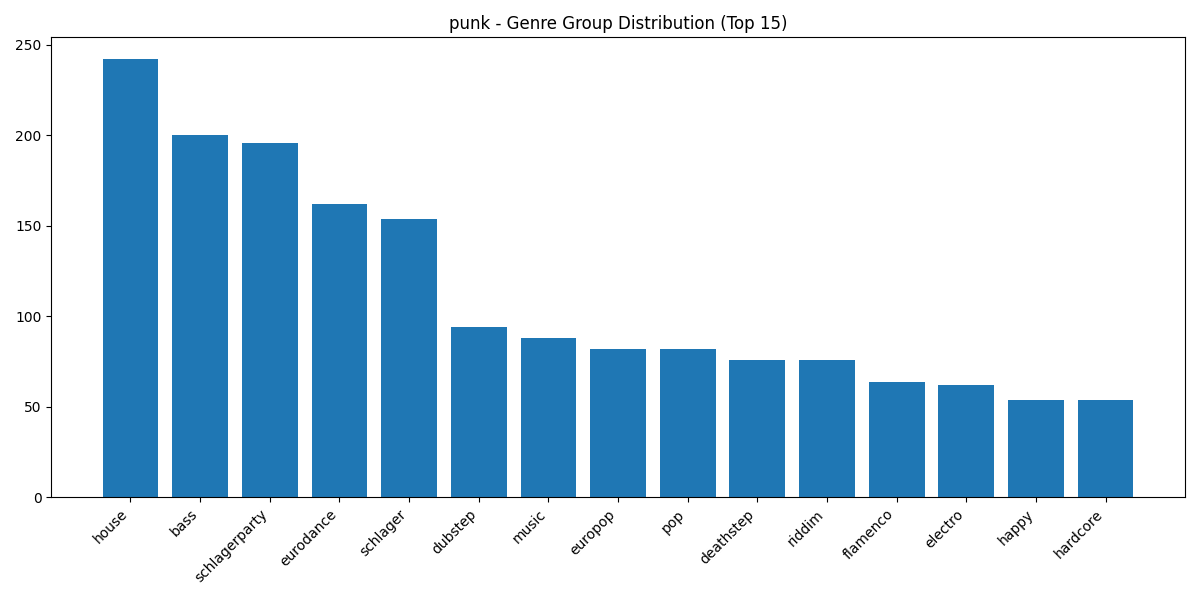}
        \caption{Punk Group Distribution}
    \end{subfigure}
    \begin{subfigure}{0.24\textwidth}
        \centering
        \includegraphics[width=\linewidth]{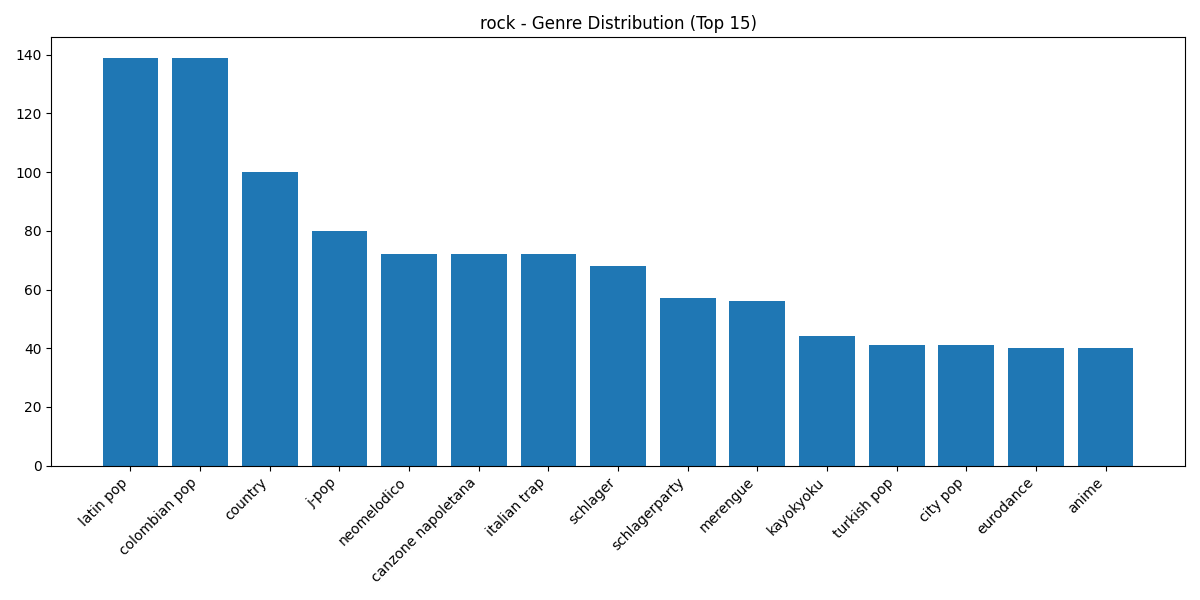}
        \caption{Rock Distribution}
    \end{subfigure}
    \begin{subfigure}{0.24\textwidth}
        \centering
        \includegraphics[width=\linewidth]{results/distributions/rock_distribution.png}
        \caption{Rock Group Distribution}
    \end{subfigure}
    
    \begin{subfigure}{0.24\textwidth}
        \centering
        \includegraphics[width=\linewidth]{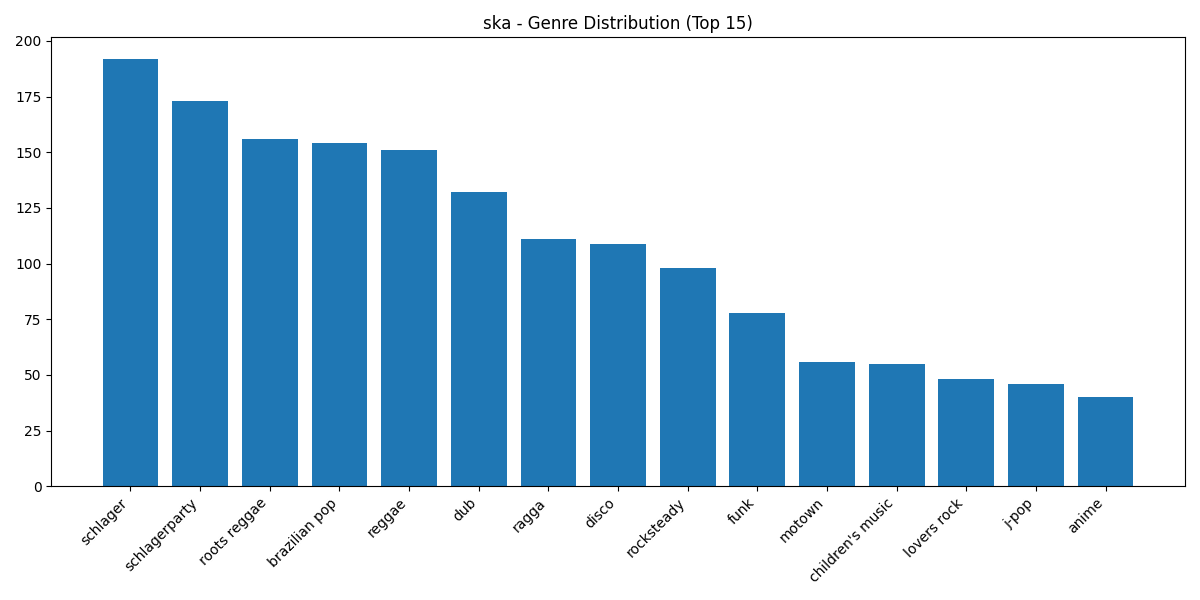}
        \caption{Ska Distribution}
    \end{subfigure}
    \begin{subfigure}{0.24\textwidth}
        \centering
        \includegraphics[width=\linewidth]{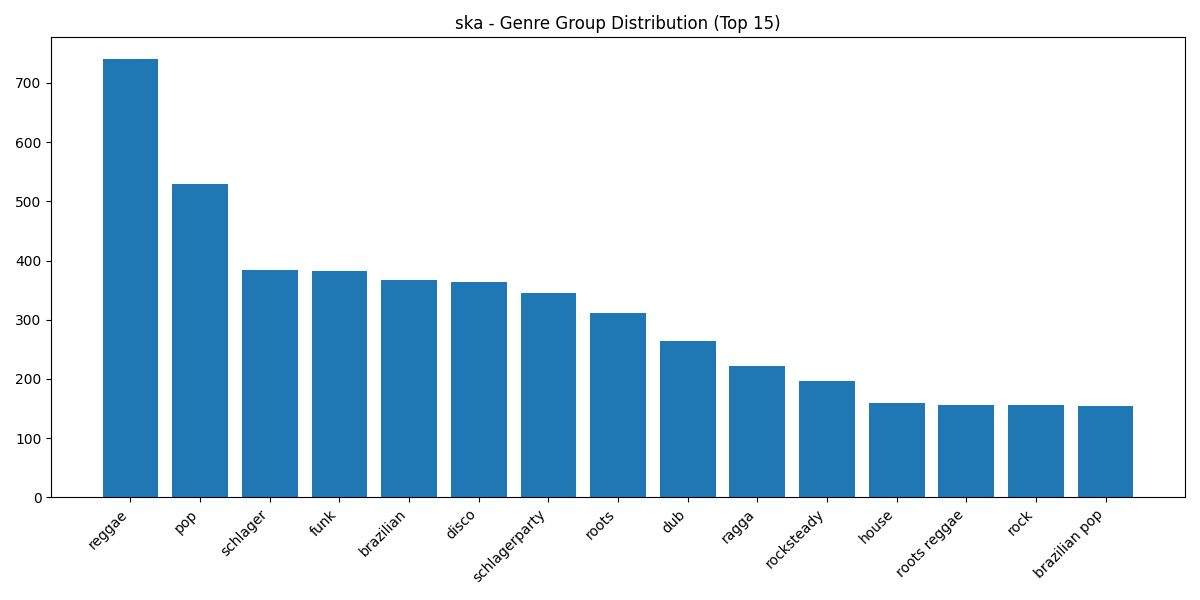}
        \caption{Ska Group Distribution}
    \end{subfigure}
    \begin{subfigure}{0.24\textwidth}
        \centering
        \includegraphics[width=\linewidth]{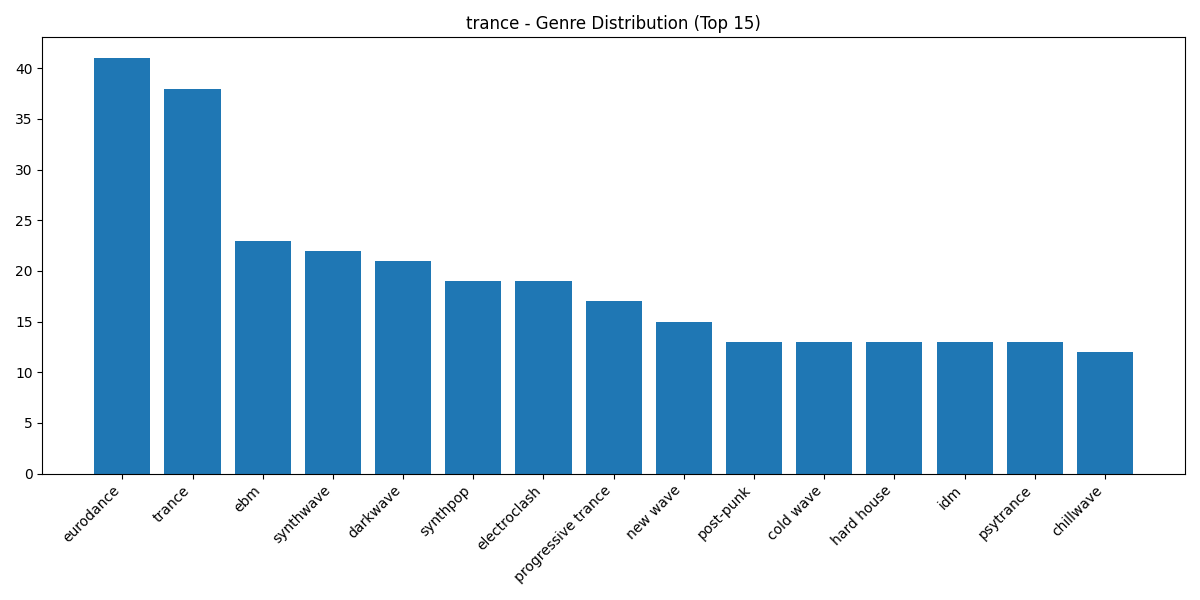}
        \caption{Trance Distribution}
    \end{subfigure}
    \begin{subfigure}{0.24\textwidth}
        \centering
        \includegraphics[width=\linewidth]{results/distributions/trance_distribution.png}
        \caption{Trance Group Distribution}
    \end{subfigure}

    \caption{Distributions of various music genres and substring groups in the ground truth ontology at the locations extracted from the LLM for the same genre as a search term.}
    \label{fig:genreDistributions}
\end{figure*}

The first accuracy test performed is the analysis of songs found in the ground truth data at the locations returned for individual query formulations per genre. For each query formulation of a given genre, up to 50 songs were sampled randomly from the high-dimensional bin at the LLM extracted location. These songs were then combined across all query formulations of the same genre. Lastly, the distribution of genres across all these songs was analyzed to visualize which genres are present in the ground truth data at the locations where the model assumes the genre's location. In addition to the raw genre strings, we also did a grouping operation that analyzed the distribution of shared substrings among genre names, which was found to reduce the effect of highly specialized genres. For each genre, a Distribution chart of both raw genre and substring occurrence across individual search locations was generated, some of which are shown in Figure \ref{fig:genreDistributions}.

We can see that for some genres, the genre searched for indeed has the highest presence (e.g., jazz, classical, pop, R\&B). For some others, a clear relation between top results and search query can be recognized, such as "merengue", "música", "salsa", etc., as the top returned genres for "latin". For other genres, however, when analyzed in this manner, the accuracy seems far off (e.g., metal, punk, rock, ska, trance).

While interesting, this type of analysis, beyond its lack of quantification, has an inherent flaw. This is because it doesn't take the global genre distribution of the ground truth data into account. This means that if a genre is globally very present in the data, it will always dominate the distribution, leading to falsely positive accuracy for common genres and falsely negative accuracy for less common genres, which can never dominate the distribution. 
This is because it might be that there is no location in the vector ontology, which has a chosen genre as the majority of its songs. Only if the correlation between the audio features used in the vector ontology and the chosen genre is high enough and little overlap with other genres exists, we can be certain that we could even find a location that returns promising results under the above analysis. 

\subsubsection{Qualitative spatial Comparisons}

To overcome the limitation of the above methodology, we created a more comprehensive analysis determining the accuracy of genre placements in the extracted LLM worldview compared to our ground truth vector ontology populated by Spotify data. Concretely, we investigate how closely the spatial genre placement extracted from the LLM correlates with the true placement of the genre. Since the genre might be present in many different locations of the ground truth vector ontology, some sort of measure of presence based on spatial location is required.

\begin{figure*} 
    \centering

    \begin{subfigure}{0.24\textwidth} 
        \centering
        \includegraphics[width=\linewidth]{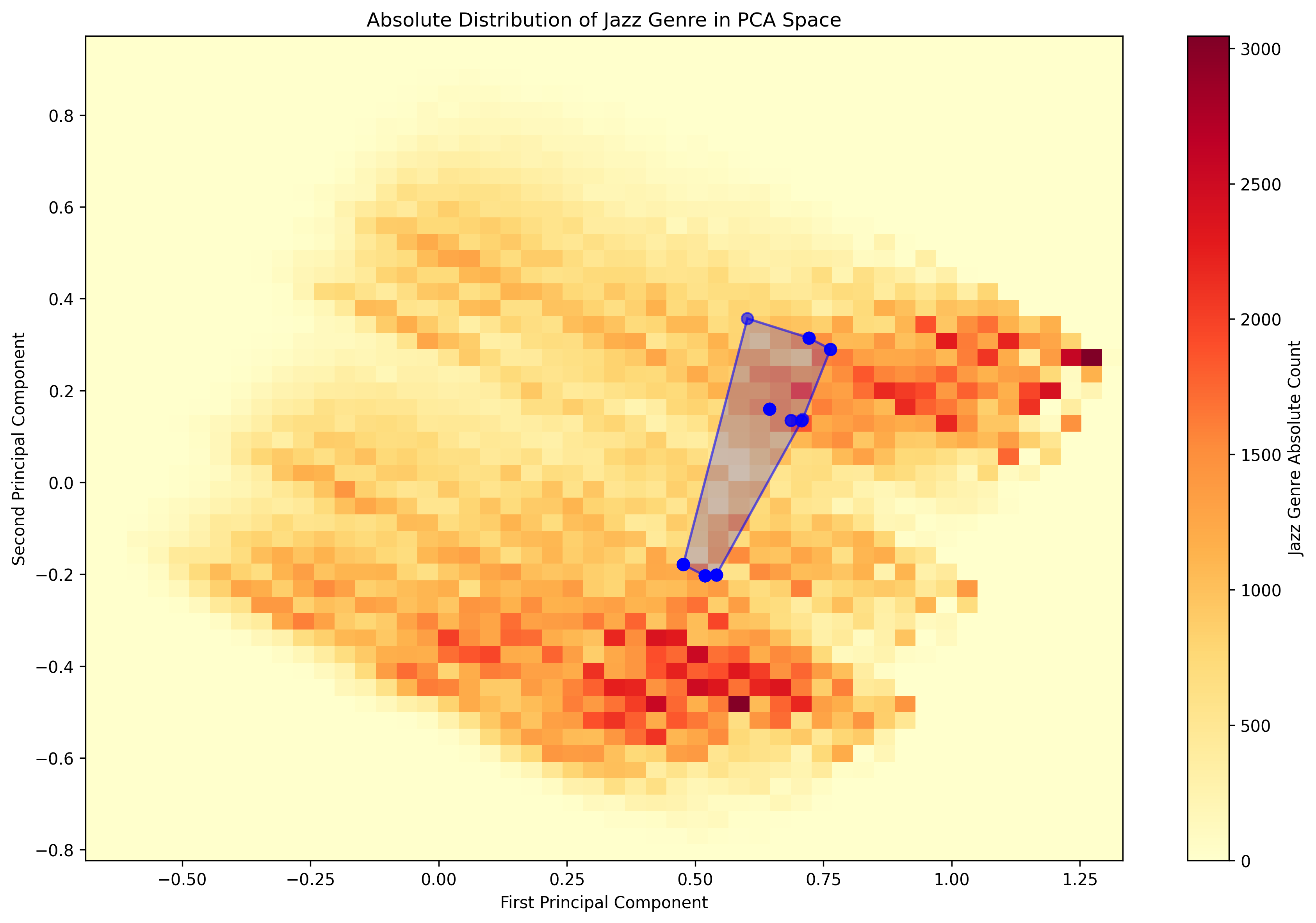}
        \caption{Jazz}
    \end{subfigure}%
    \begin{subfigure}{0.24\textwidth}
        \centering
        \includegraphics[width=\linewidth]{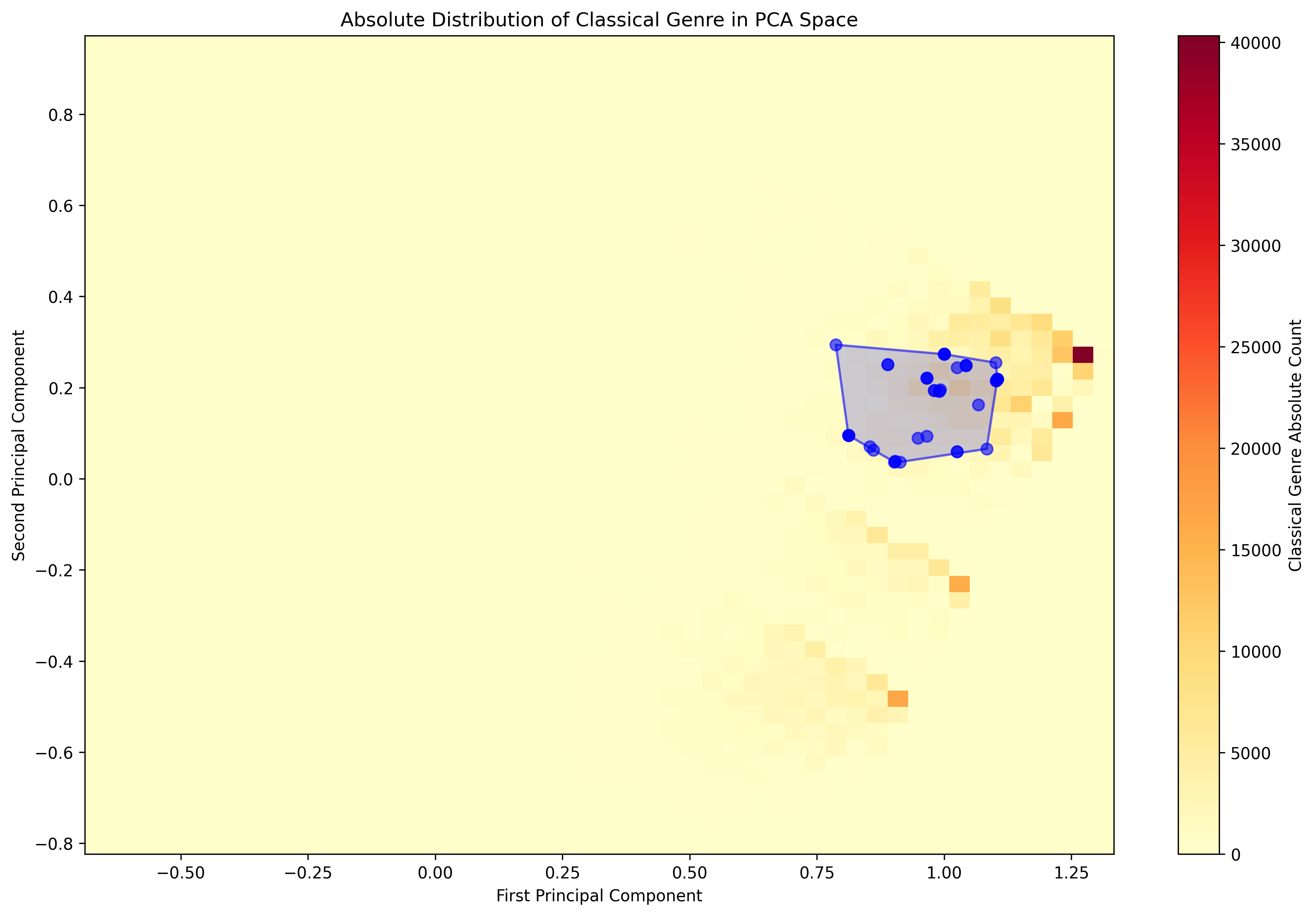}
        \caption{Classical}
    \end{subfigure}%
    \begin{subfigure}{0.24\textwidth}
        \centering
        \includegraphics[width=\linewidth]{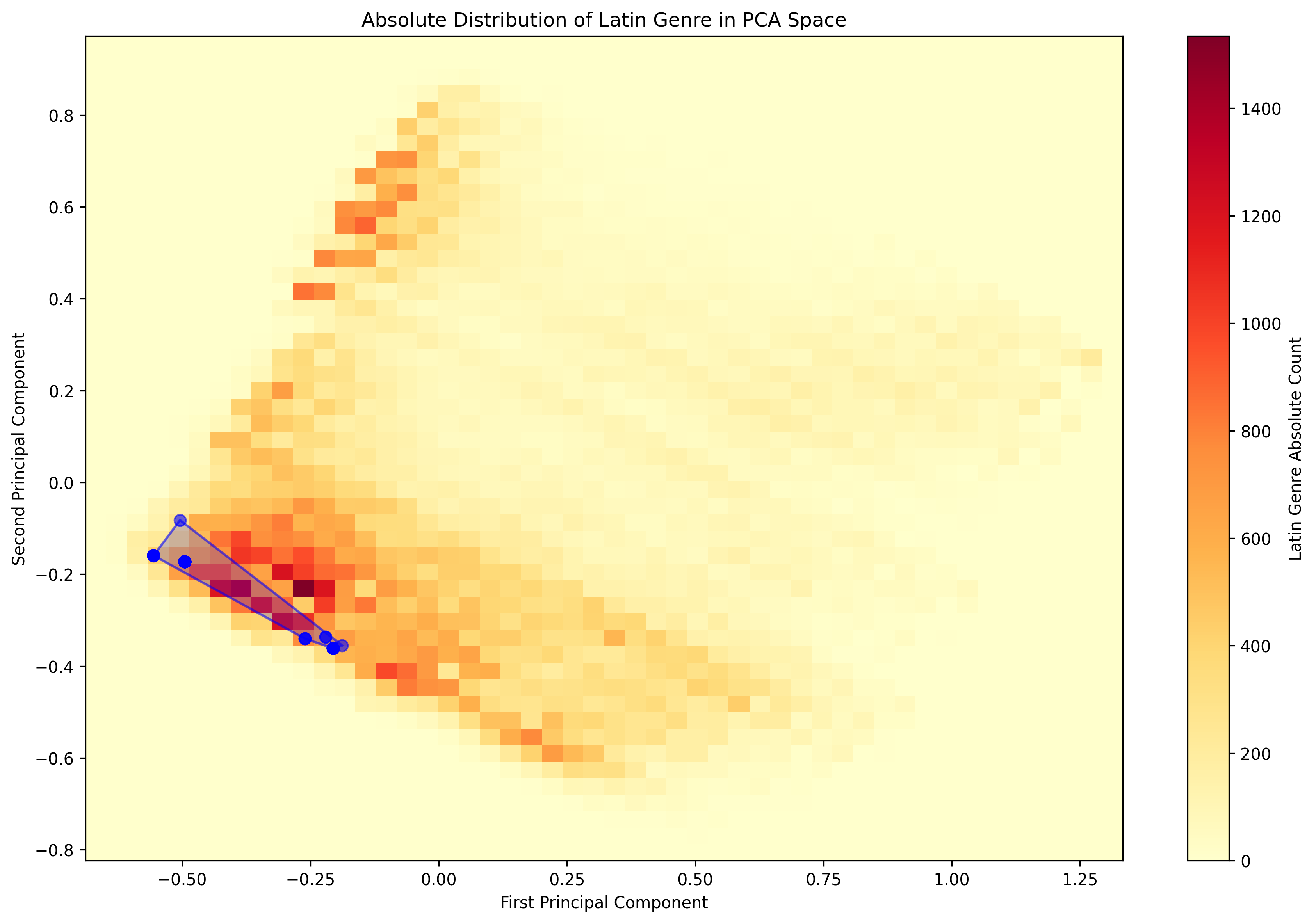}
        \caption{Latin}
    \end{subfigure}%
    \begin{subfigure}{0.24\textwidth}
        \centering
        \includegraphics[width=\linewidth]{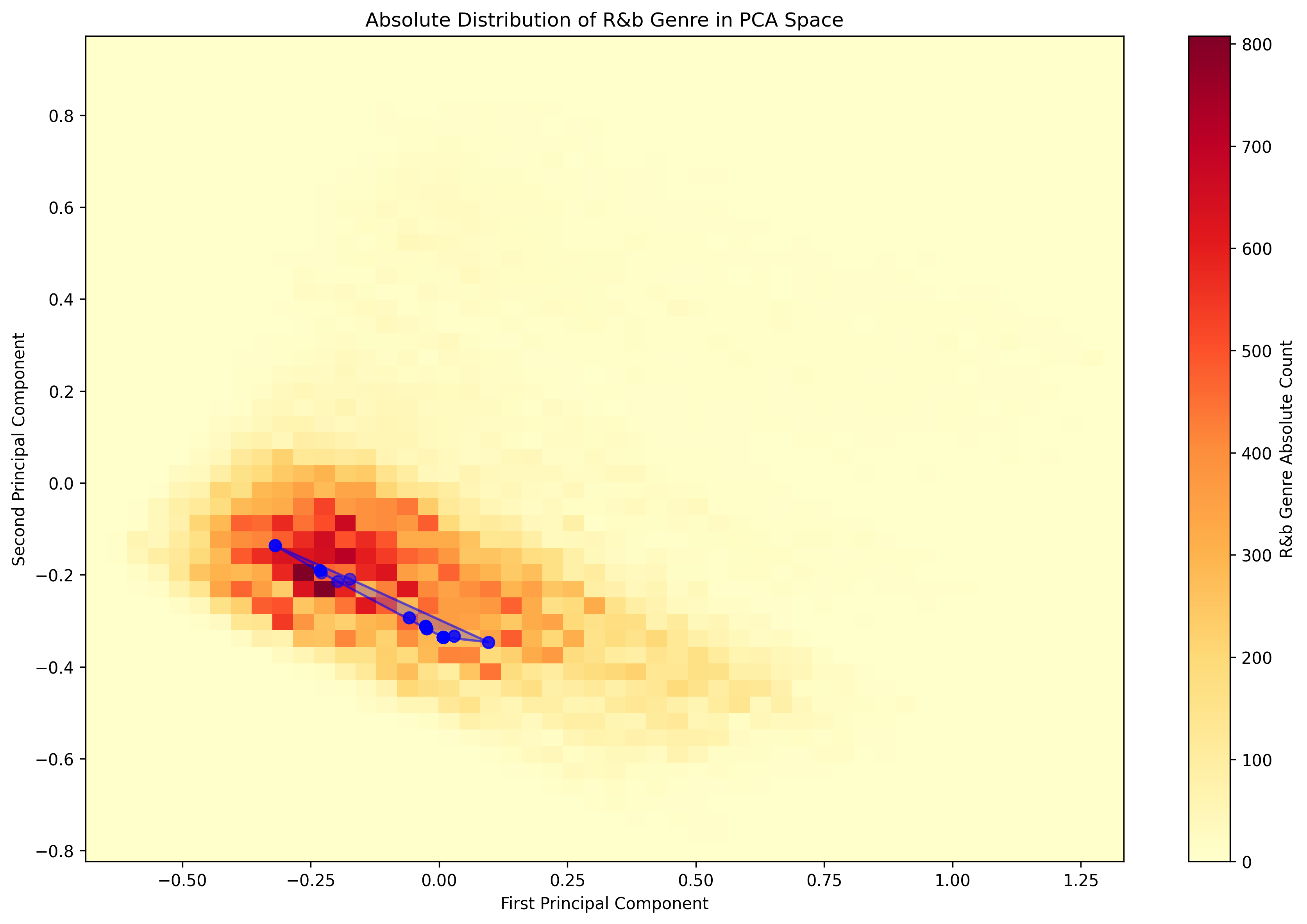}
        \caption{R\&B}
    \end{subfigure}

    \begin{subfigure}{0.24\textwidth}
        \centering
        \includegraphics[width=\linewidth]{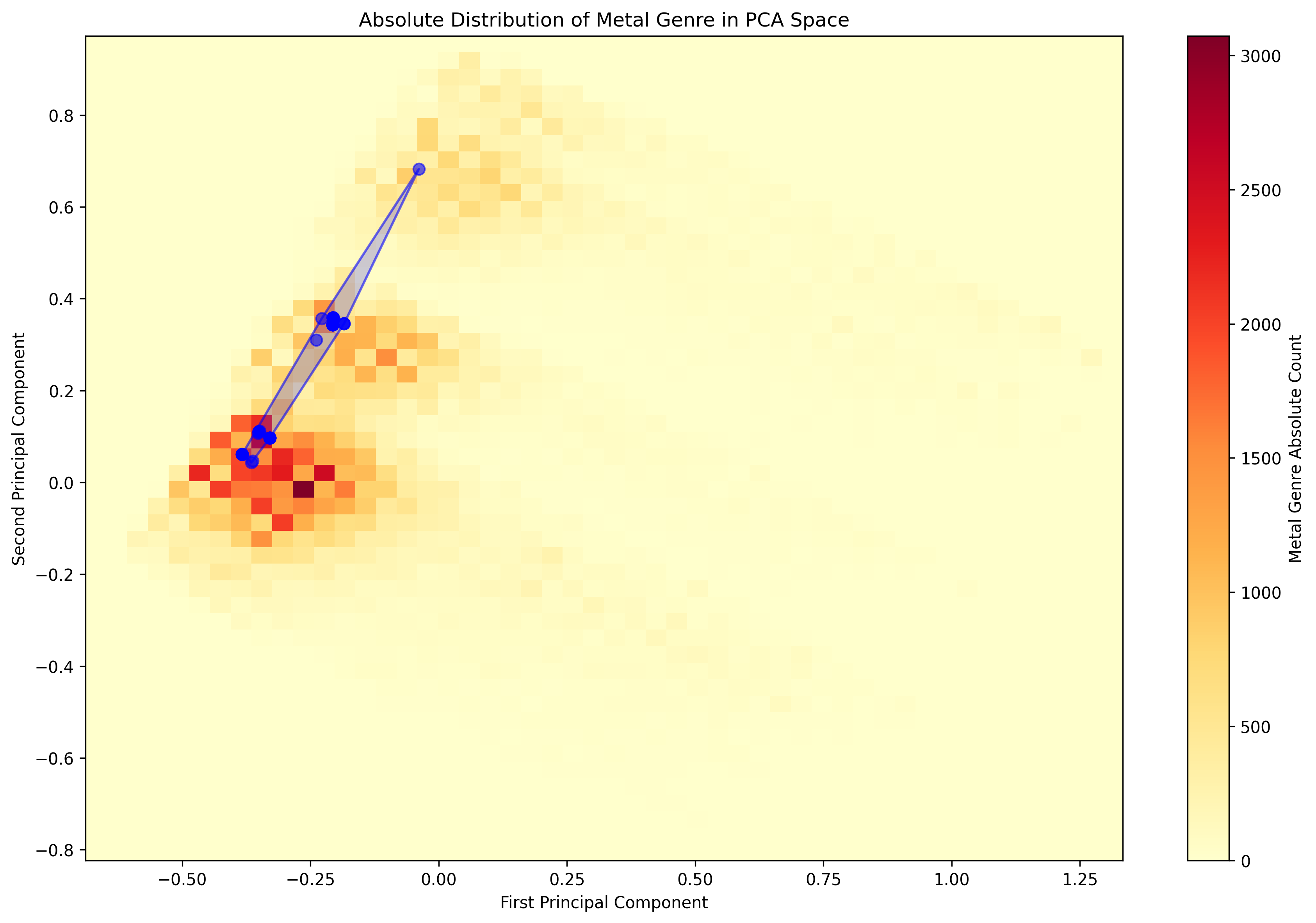}
        \caption{Metal}
    \end{subfigure}%
    \begin{subfigure}{0.24\textwidth}
        \centering
        \includegraphics[width=\linewidth]{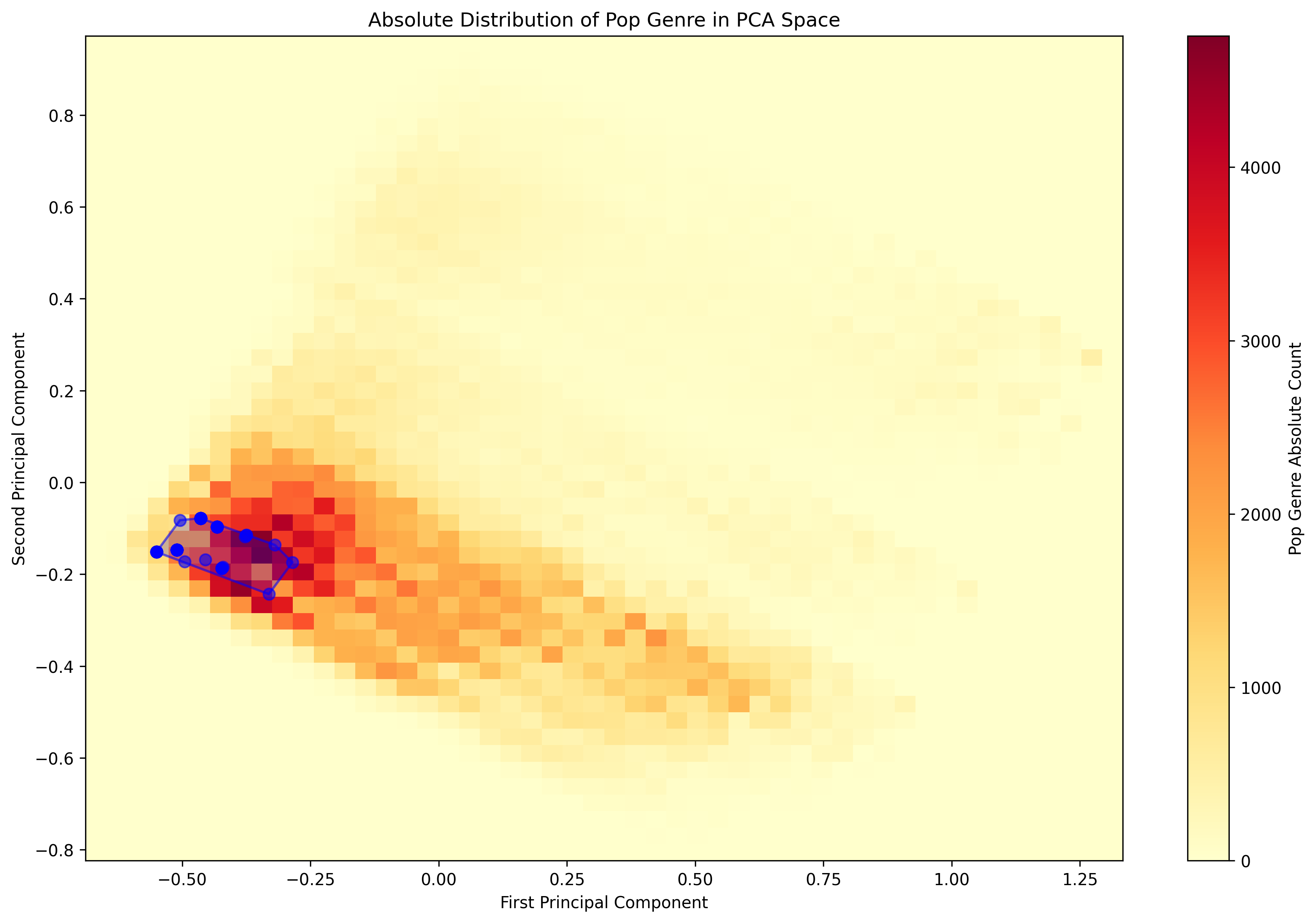}
        \caption{Pop}
    \end{subfigure}%
    \begin{subfigure}{0.24\textwidth}
        \centering
        \includegraphics[width=\linewidth]{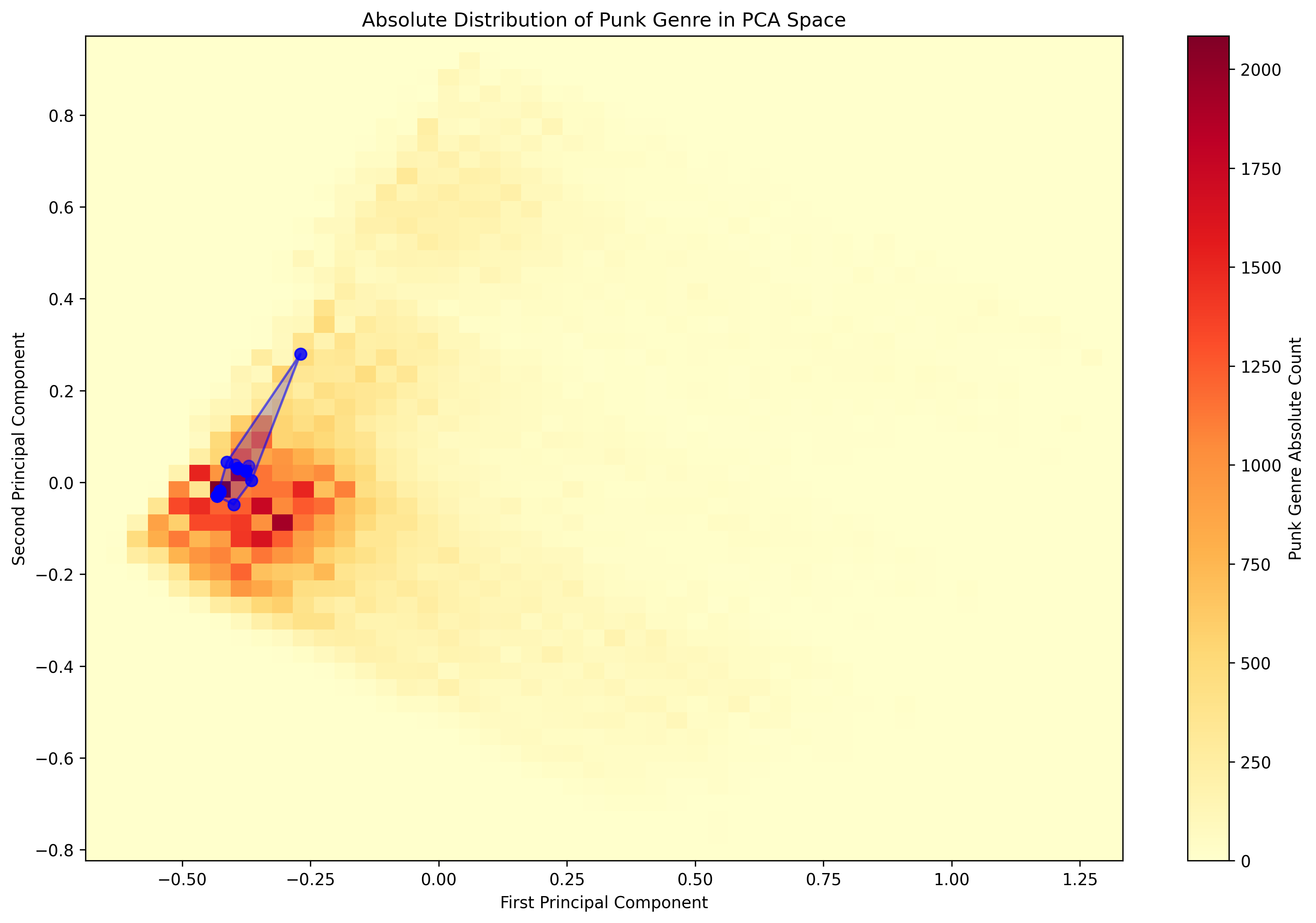}
        \caption{Punk}
    \end{subfigure}%
    \begin{subfigure}{0.24\textwidth}
        \centering
        \includegraphics[width=\linewidth]{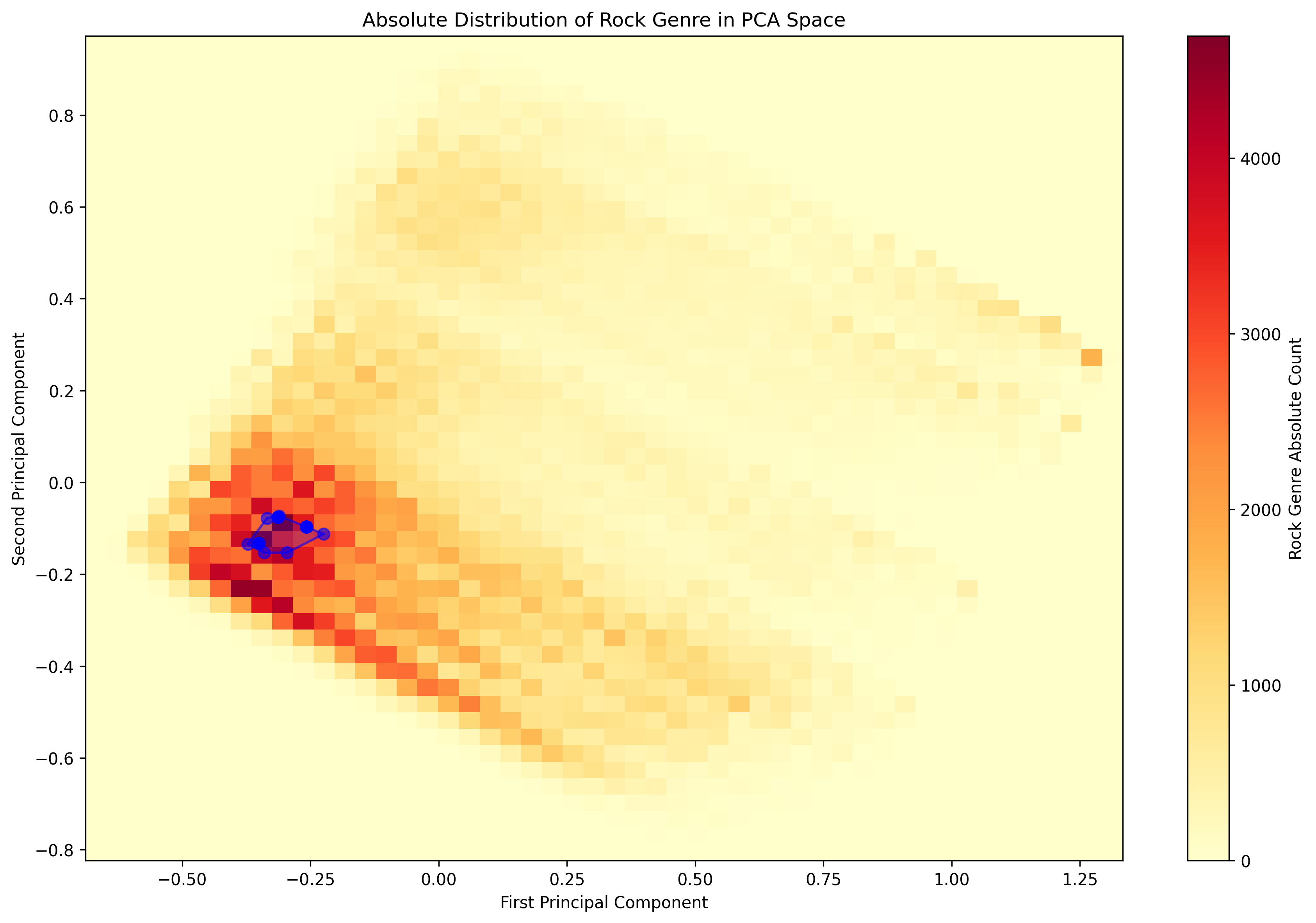}
        \caption{Rock}
    \end{subfigure}

    \begin{subfigure}{0.24\textwidth}
        \centering
        \includegraphics[width=\linewidth]{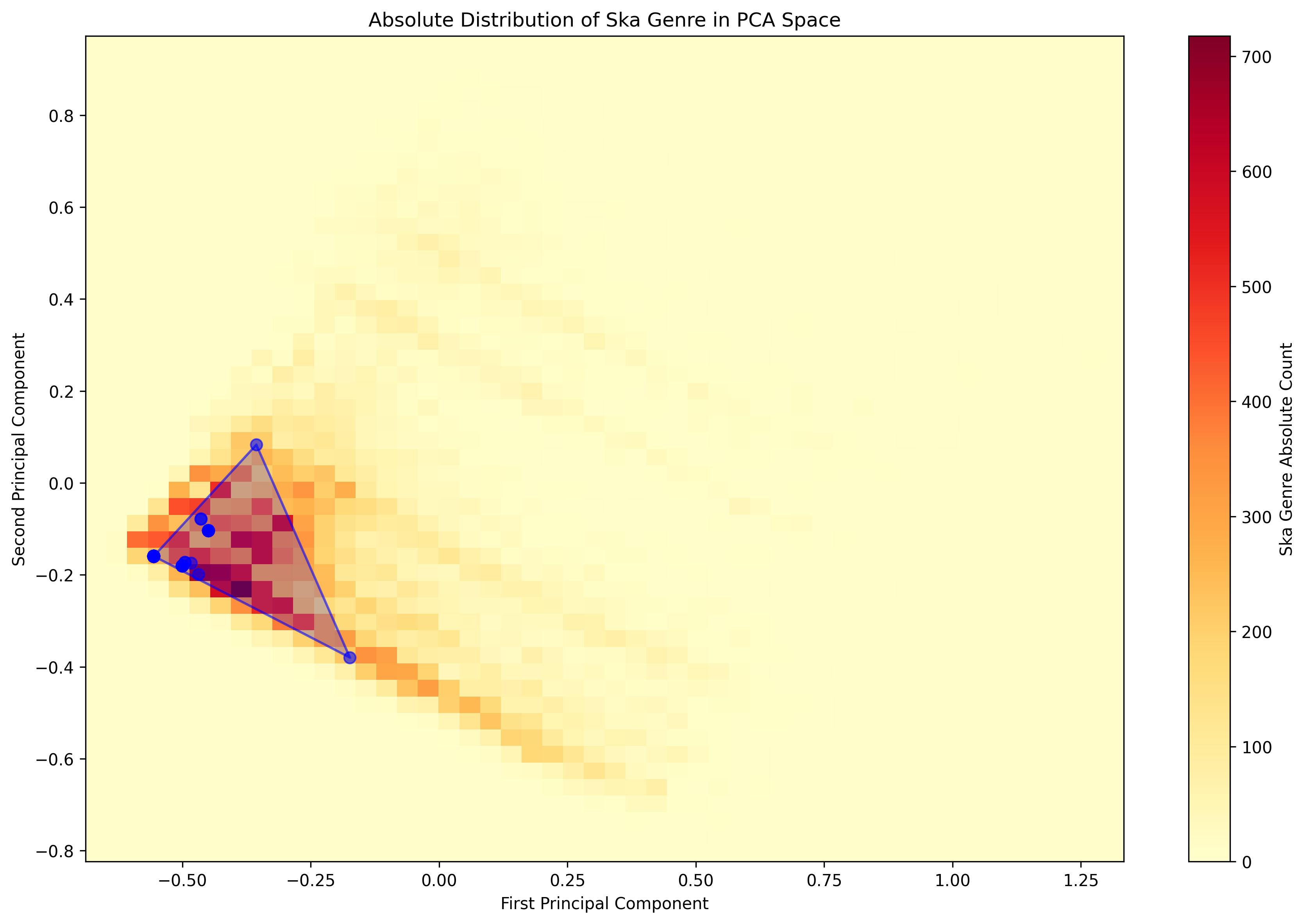}
        \caption{Ska}
    \end{subfigure}%
    \begin{subfigure}{0.24\textwidth}
        \centering
        \includegraphics[width=\linewidth]{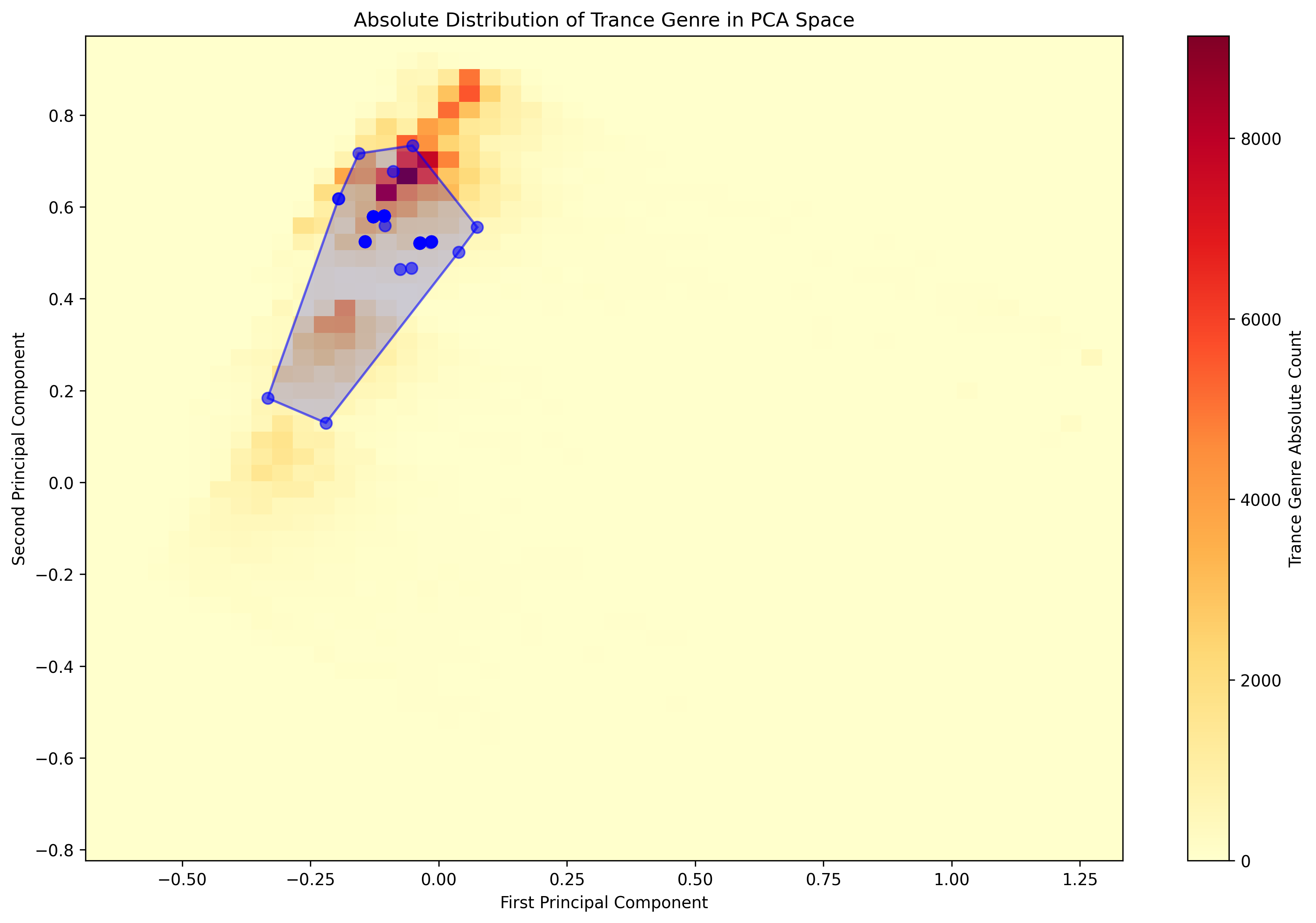}
        \caption{Trance}
    \end{subfigure}%

    \caption{The true distribution of genre occurrence in the ground truth vector ontology (heatmap), overlayed by the LLM extracted genre locations (blue) for an identical ontology, both projected using PCA from section 6.2}
    \label{fig:genreHeatmaps}
\end{figure*}

To analyse this qualitatively, we generated a heat map of presence for each genre of interest. This was done by counting the number of songs belonging to the genre in any given hypercube bin and projecting the bin's center point onto the PCA generated in section 6.2. The heat map visualizes the average genre count per bin location projected onto the given heatmap grid cell. Figure \ref{fig:genreHeatmaps} shows the heatmaps for selected genres (same selection as in 6.3) overlayed by the individual search points extracted from the LLM for the same genre across query variations. This gives a direct spatial comparison between the LLM's perception of genre placement and its true occurrence in the ground truth ontology. 

Figure \ref{fig:genreHeatmaps} demonstrates that the LLM extracted vector ontology closely overlaps with the ground truth ontology. For example, for the genre metal, the extracted worldview aligns with the ground truth at incredible resolution, capturing both the location and importance of three different clusters close to perfectly. Comparing the heatmaps to the distribution analysis in the previous section further shows the weakness of the distribution analysis, as genres that indicated lack of accuracy in their distribution show close alignment (e.g. Metal, Punk, Rock, Ska, Trance) while others seem less precise despite an indication from the distribution data (e.g. Jazz, Classical). This is due to the uneven genre distribution within the ground truth data.

\subsubsection{Quantitative Comparison}

To quantitatively measure the accuracy of the extracted worldview, we compare the centroids of LLM-extracted genre positions with the centroid of the same genre in the ground truth ontology. To create said centroid for the ground truth data, a weighted average was created between all locations in the ontology that contained the genre, weighted by the number of songs of said genre present. Three measures for accuracy were created. Firstly, the Euclidean distance between LLM extracted genre centroids and ground truth centroids was measured. Secondly, the cosine similarity was measured. Lastly, the cosine similarity in a shifter space was measured where the origin was moved to the center of the space (i.e., translating the space from 0-1 to -0.5 - 0.5), since otherwise all points lie in the same quadrant, creating inherent cosine similarity. All three measures were compared against a random baseline, which randomly sampled from both search centroids and ground truth centroids, creating the same measures.

\begin{figure}
    \includegraphics[width=\linewidth]{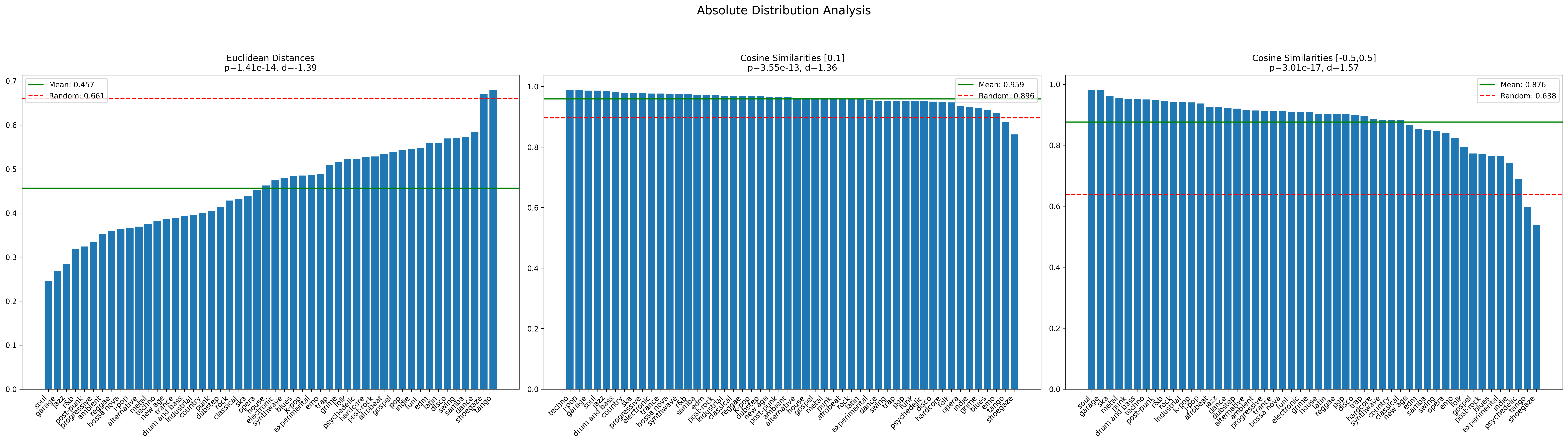}
    \caption{Alignment between genre centroids in the LLM extracted and ground truth ontology}
    \label{fig:accuracyQuantitative}
\end{figure}

Figure \ref{fig:accuracyQuantitative} clearly shows strong alignment on all three measures. The mean Euclidean distance is 0.46 compared to a random baseline of 0.66 with high significance (p \textless 0.000000, d = -1.39). The Cosine Similarity in original space has a mean of 0.96 compared to 0.9 random (significance (p \textless 0.000000, d = 1.36) and in the adjusted space, the improvement is even more visible with 0.88 compared to 0.64 random (p \textless 0.000000, d = 1.57). Overall, this Clearly indicates that the vector ontology extracted fromt the LLM is strongly aligned with the ground ontology.

\subsection{Additional Analysis}

\begin{figure}
\centering
    \includegraphics[width=0.9\linewidth]{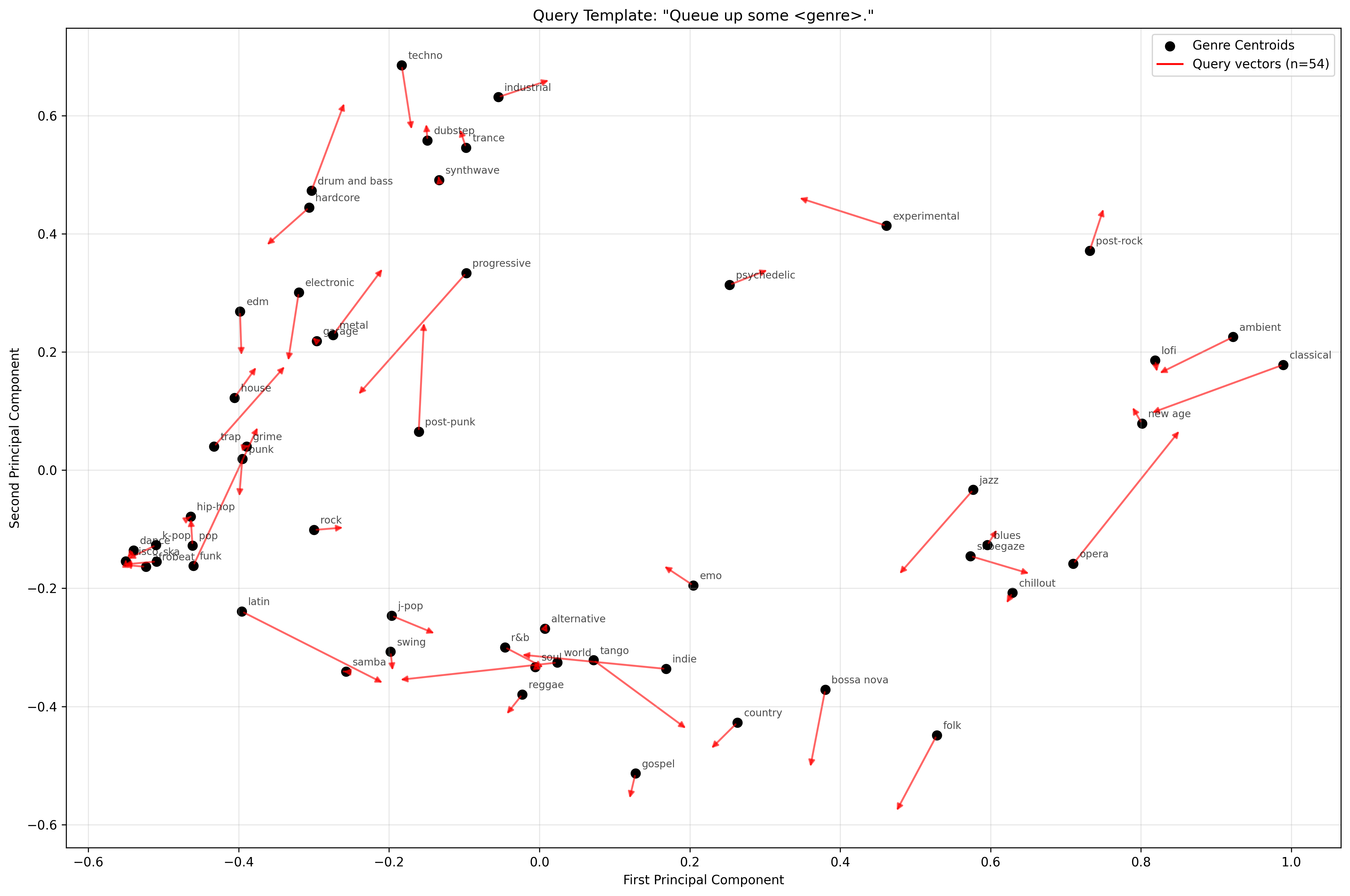}
    \caption{Cross formulation search centers (black) vs. query search location for formulation: "Queue up some \{genre\}" (red arrowhead), projected using the PCA projection generated in section 5}
    \label{fig:queryDirections}
\end{figure}

The Consistency Analysis, even though indicating overall consistency, still raises a question about the origin of the variance. Assuming that the used methodology truly extracts a learned feature space created during pretraining, it would require perfect consistency as a deterministic projection. However, the use of different query formulations leads to a change in internal latent space representation, which consequently explains output variance. We hence analyze the connection between individual query formulations and hence latent space representation and the location shifts in our vector ontology. We argue that a global correlation between input formulation and output variance in the predefined ontology would strengthen our hypothesis for the link between the learned latent space and our vector ontology. \\
We hence analyze the similarity between the vectors connecting the search centroids of each genre to the individual search location for a specific query formulation. For example, Fig \ref{fig:queryDirections} shows these vectors for all genres and the query formulation "Queue up some \{genre\}".

\begin{figure}
\centering
    \includegraphics[width=\linewidth]{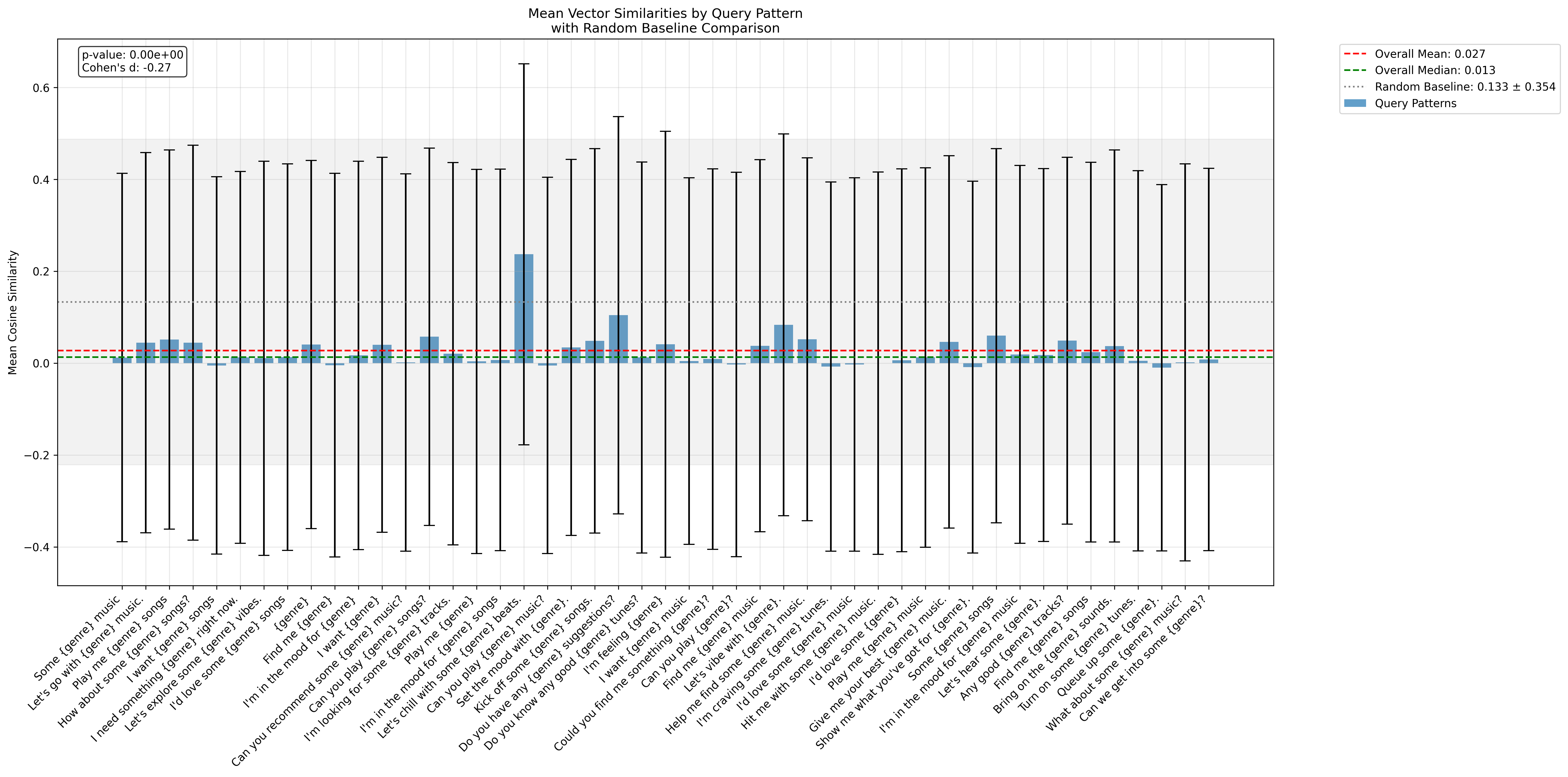}
    \caption{cosine similarity between cross-genre centroid-formulation vector pairs with identical query formulation}
        \label{fig:globalQueryCorrelation}
\end{figure}

Figure \ref{fig:globalQueryCorrelation} shows the cosine similarity between pairs of such vectors for each query formulation. We can see that there is no significant correlation with a mean of 0.013 compared to a random 0.13.  Only one formulation shows a consistent shift in our vector ontology, namely "Let's chill with some \{genre\} music". This seems somewhat logical as it is the only query that mentions a clear emotion or mood, in this case "chill" which reasonably could create a global shift in, for example, the energy dimension. \\
When inspecting Figure \ref{fig:queryDirections} closely, however, we notice that while not globally ordered, the arrows also seem not to be fully chaotic. More concretely, we can see local consistency (e.g., between country, bossa nova, and folk, or between dubstep and trance). We hence created a second cosine similarity analysis which tests the cosine similarity of not all genres, but only a genre and its 5 nearest neighbours. This, so we hypothesize, can uncover local consistency in query formulation shifts from genre centroids.

\begin{figure}
\centering
    \includegraphics[width=\linewidth]{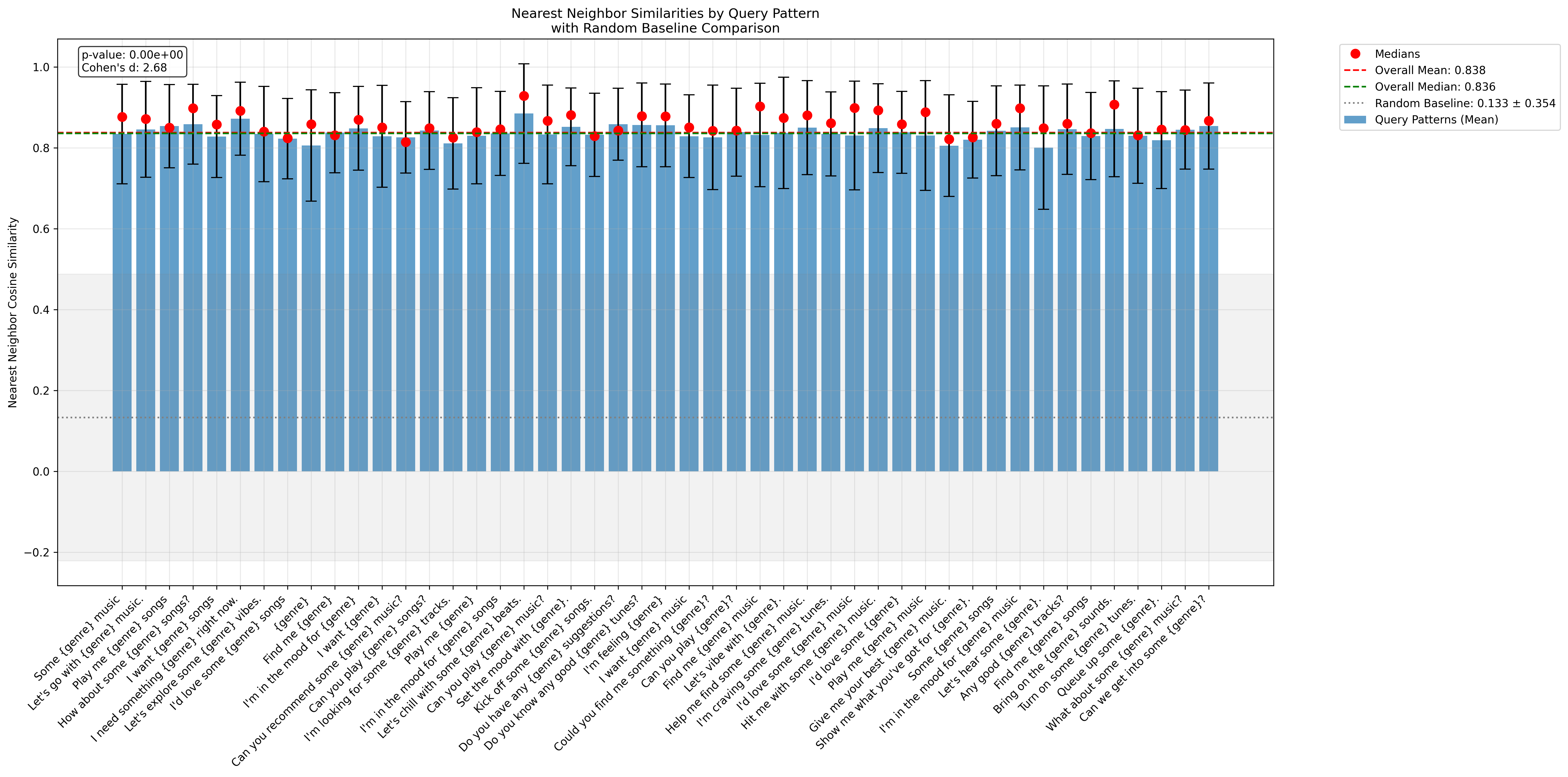}
    \caption{5-nearest-neighbors cosine similarity of cross-genre centroid-formulation vector pairs with identical query formulation}
        \label{fig:localQueryCorrelation}
\end{figure}
Figure \ref{fig:localQueryCorrelation} indeed shows a surprising local consistency of position shifts based on query formulation with a mean correlation of 0.84 (random 0.133) and high statistical significance (p \textless 0.00, d = 2.68). The local bound of this interaction is most probably rooted in context dependence. We find this somewhat logical as "vibing with \{genre\}" might indeed have a different connotation in the case of classical music than in the case of dubstep.\\

Regardless of its exact origin, the correlation shown in Figure \ref{fig:localQueryCorrelation} seems specifically interesting to us, since it draws a direct correlation between latent space representation (query formulation) and position in the vector ontology further strengthening evidence for the vector ontology having a direct link to the latent space. 

\section{Discussion}

The results of our analysis provide strong evidence supporting both of our initial hypotheses and strengthening the link between vector ontologies and AI latent space representations as proposed in  \parencite{Paper0}.

Firstly, we demonstrated in Section 6 that Large Language Models do indeed possess a consistent and extractable internal world model that can be projected onto a predefined vector ontology. The high spatial consistency of genre placements across several measures, and the clear clustering patterns visible in our PCA visualizations indicate that the model maintains a stable and structured understanding of musical relationships within our defined feature space. This indicates that during its pertaining, the model learned to represent musical relationships in a way that is consistent and extractable using a predefined vector ontology.

Second, our accuracy analysis confirms that this extracted world model significantly correlates with the true distribution of musical features in our dataset. The alignment between search locations and actual genre distributions suggests that the LLM has internalized meaningful relationships between musical characteristics during its training process, which are aligned with real-world data. This alignment is especially noteworthy given that the model was never explicitly trained on our specific vector ontology or musical feature space, to our knowledge.

\subsection{Key Observations}
\begin{itemize}
\item \textbf{Genre-Dependent Performance}: The model's accuracy varies considerably across different genres. This pattern might reflect the clarity and consistency of genre definitions in the training data, as well as varying correlation between audio features and genre.

\item \textbf{Dimensional Utilization}: The varying dimensionality of genre representations (mean of 6.08 dimensions) suggests that the model knew to discriminate genres using different subsets of our feature space. This efficient use of dimensions aligns with musical intuition – not all audio features are equally relevant for distinguishing every genre.

\item \textbf{Spatial Organization}: The overlap patterns between genre regions in our PCA visualizations reflect meaningful musical relationships (e.g., jazz-blues overlap, classical-opera proximity), indicating that the model has captured nuanced musical similarities beyond simple genre categorizations.
\end{itemize}

\subsection{Implications}
\begin{itemize}
\item \textbf{World Model Extraction}: Our methodology demonstrates that it's possible to extract and utilize the structured world model embedded within LLMs in a way that maintains interpretability and allows for verification against ground truth data. This is a crucial step towards the development of more interpretable and verifiable LLMs and knowledge extraction systems.

\item \textbf{Implications for Information Retrieval}: The strong alignment between the model's projections and actual data distributions, as well as model consistency, suggests that this approach could be effective for retrieval systems, particularly in domains where relationships between entities can be meaningfully represented in vector space. The presented method already represents such a retrieval system as it retrieves songs from the vector ontology based on the model's internal world model and a given user query.

\item \textbf{Query formulation consistency}: The local consistency of shifts in the vector ontology based on query formulation indicates a direct link between latent space and vector ontologies. If this holds up, it has important implications, but it requires further investigation.

\end{itemize}
\subsection{Limitations}
Several limitations should be noted. First, our vector ontology, while comprehensive, may not capture all relevant musical features. Second, the reliance on Spotify's genre tags as ground truth introduces some uncertainty, as these tags are not always consistent or complete. Finally, our analysis focuses on a single domain (music) and may not generalize equally well to all knowledge domains. Specifically, the creation of vector ontologies for other domains is a challenging task, and most domains do not have predefined and pre-scored data spanning a high-dimensional, ontologically meaningful vector space.\\
Most importantly, the current methodology extracts the world model from the LLM through the generation of tokens and queries asking for its internal representation of the world. This is a very indirect way of accessing the world model and might not fully capture the true world model of the LLM; accordingly, methods should be developed to extract the world model directly from the LLM's dimensional latent space. It is however, reasonable to believe that some form of token prediction is inherently necessary, as it is the task these models are trained for. Therefore, its internal world model is most likely not contained in a single forward pass, but rather in the recursive application of predicted tokens.

\section{Conclusion}

Our results support our hypothesis that LLMs contain extractable, structured knowledge that can be mapped onto interpretable frameworks, showcasing meaningful relationships with real-world data distributions.\\
We further show that, specifically, the projection of an LLM's internal world model onto vector ontologies as proposed in our earlier work \parencite{Paper0} is a promising method to extract and verify its understanding of the world. We provide clear evidence for both the consistency and accuracy of the extracted worldview and showcase initial evidence for a clear connection between latent space representations of LLMs and vector ontologies.

\setlength{\bibitemsep}{10pt}   
\printbibliography

\end{document}